\documentclass[11pt]{article}
\usepackage[final]{acl}
\usepackage{pgfplots}
\pgfplotsset{compat=1.18}
\usepgfplotslibrary{colormaps,groupplots}

\usepackage{times}
\usepackage{latexsym}
\usepackage[T1]{fontenc}
\usepackage[utf8]{inputenc}
\usepackage{microtype}
\usepackage{inconsolata}
\usepackage{graphicx}
\usepackage{subcaption}
\usepackage{xcolor}
\usepackage[most]{tcolorbox}
\usepackage{amsmath}
\usepackage{amssymb}
\usepackage{amsthm}
\usepackage{algorithm}
\usepackage{algpseudocode}
\usepackage{booktabs}
\usepackage{multirow}

\newtcolorbox{observationbox}{
  colback=blue!8,
  colframe=blue!45!black,
  boxrule=0.6pt,
  arc=2mm,
  left=1mm,
  right=1mm,
  top=0.6mm,
  bottom=0.6mm
}
\theoremstyle{definition}
\newtheorem{theorem}{Theorem}
\title{VoidPadding: Let \texttt{[VOID]} Handle Padding in Masked Diffusion Language Models
    so that \texttt{[EOS]} Can Focus on Semantic Termination}

  \author{
  \textbf{Chunyu Liu\textsuperscript{1}},
  \textbf{Zhengyang Fan\textsuperscript{1}},
  \textbf{Kaisen Yang\textsuperscript{1}},
  \textbf{Alex Lamb\textsuperscript{1}}\thanks{Corresponding author.}
  \\
  \textsuperscript{1}Tsinghua University
  \\
  \small{\texttt{\{liucy24,fzy24,yks23\}@mails.tsinghua.edu.cn}}
  \\
  \small{\texttt{lambalex@tsinghua.edu.cn}}
  }

\begin{document}
\maketitle
\begin{abstract}
MDLMs generate text by denoising a preallocated masked response canvas, making response-length modeling central to instruction tuning.
 Existing MDLMs often inherit the autoregressive convention of using repeated
  \texttt{[EOS]} tokens for padding during instruction tuning, giving \texttt{[EOS]} a dual role as both a
  semantic terminator and a padding token. We show that this dual role is a root
  cause of \texttt{[EOS]} overflow under large-block decoding.
 To decouple these roles, we propose VoidPadding, which introduces \texttt{[VOID]}
  for padding and reserves \texttt{[EOS]} for termination. During inference, the learned \texttt{[EOS]} signal enables early stopping, while the learned
  \texttt{[VOID]} signal guides adaptive response canvas expansion.
On Dream-7B-Instruct, VoidPadding improves the block-size-averaged four-task mean across mathematical reasoning and code generation benchmarks by \(+17.84\) points over the original model and \(+6.95\) points over RainbowPadding, while reducing decoding NFE by 55.7\% on average.
Code is available at \url{https://github.com/Haru-LCY/VoidPadding}.
\end{abstract}

\section{Introduction}
Masked diffusion language models~\cite{austin2021structured,wang2025revolutionizing,lou2023discrete,liu2025wedlm} (MDLMs) offer a promising alternative to autoregressive language models~\cite{radford2018improving} (ARLMs). Whereas ARLMs generate tokens sequentially from left to right, MDLMs use bidirectional attention to denoise the entire sequence in parallel~\cite{bie2026llada2}. This bidirectional conditioning, however, makes MDLMs sensitive to the padding tokens introduced in instruction tuning.

  During instruction tuning~\cite{chung2024scaling}, instruction--response pairs are typically processed in batches for efficiency and padded to the
  maximum sequence length within each mini-batch. 
Following a convention inherited from ARLM training, MDLMs often use repeated \texttt{[EOS]} tokens as
  padding~\cite{nie2026large}. In ARLMs, causal
  masking~\citep{vaswani2017attention} prevents earlier positions from conditioning on
  \texttt{[EOS]} padding, so \texttt{[EOS]} primarily serves as a
  semantic terminator. In MDLMs, however, \texttt{[EOS]} padding is also included in the
  training target. Thus, \texttt{[EOS]} takes on two
  roles: the semantic terminator role and the padding role.

During inference, MDLMs generate on a predefined response canvas whose positions are jointly modeled under bidirectional
  attention~\cite{cheng2025sdar}. 
  The padding role of \texttt{[EOS]} introduced during instruction tuning therefore offers a length signal: high \texttt{[EOS]} confidence
  near the tail indicates that the current canvas may exceed the desired output length~\cite{li2025beyond,yang2026rho}. However, this signal
  becomes fragile under large-block decoding. 
  Once high-confidence tail \texttt{[EOS]} tokens are generated, they become part of the denoising context and can further increase nearby
  \texttt{[EOS]} probabilities, leading to short effective responses and repeated \texttt{[EOS]}.
This failure mode is
  known as \texttt{[EOS]} overflow~\citep{kim2025rainbow}.

\begin{figure*}[t]
\centering
\includegraphics[width=\textwidth]{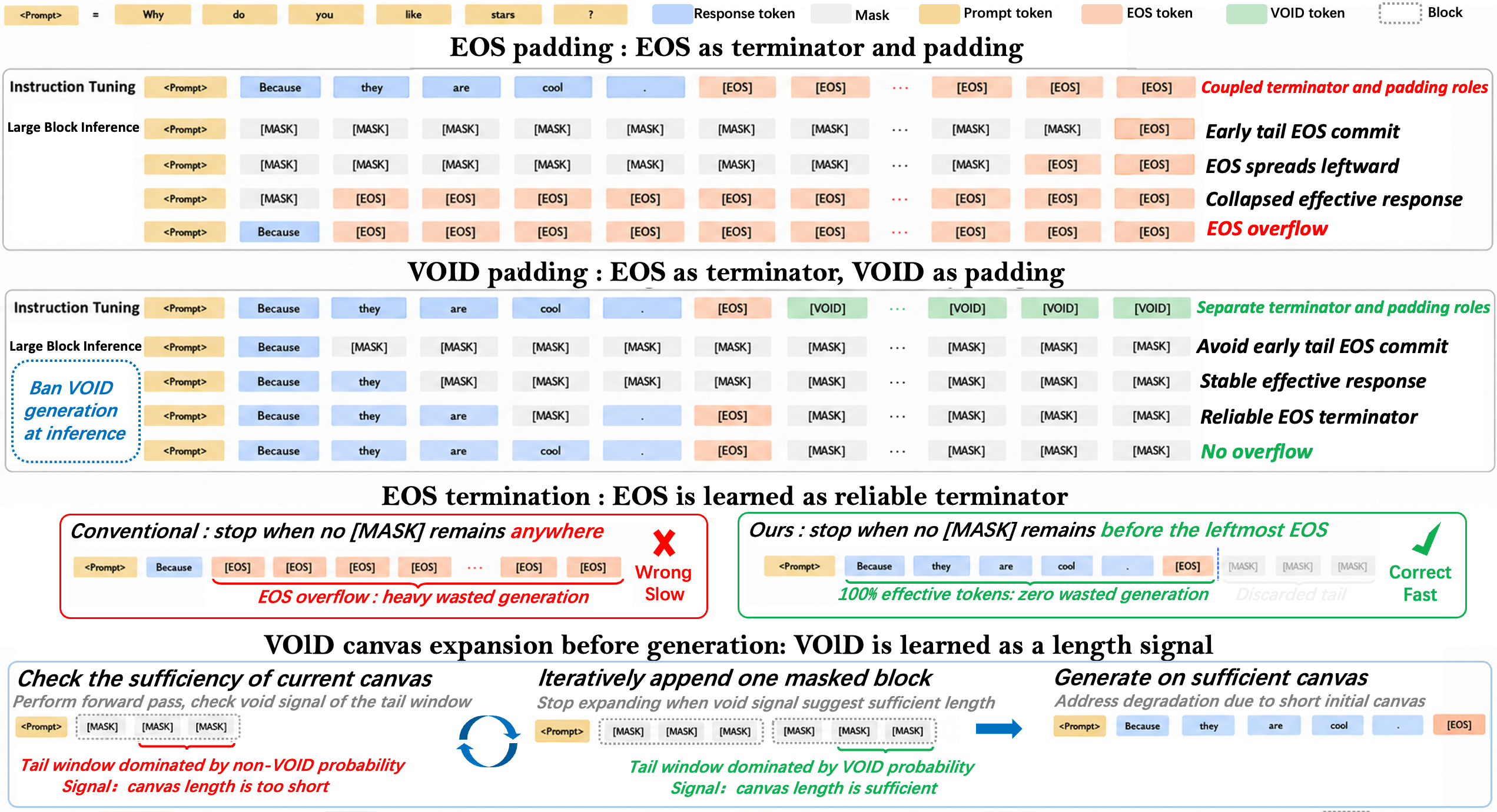}
\caption{Overview of VoidPadding. VoidPadding decouples termination from padding
  by using \texttt{[EOS]} only for stopping and \texttt{[VOID]} only for padding.
   During inference, \texttt{[VOID]} token generation is banned, decoding stops at the leftmost visible
  \texttt{[EOS]}, and the canvas may be expanded based on tail \texttt{[VOID]} signals.}
\label{fig:void-padding-overview}
\end{figure*}

 Based on these insights, we propose \textbf{VoidPadding}, a lightweight
  training--inference co-design that introduces \texttt{[VOID]} as a new padding
  token and reserves \texttt{[EOS]} solely for semantic termination.
  Training sequences are formatted as \texttt{[prompt] [response] [EOS] [VOID]*},
  where each example contains a single \texttt{[EOS]} followed by
  \texttt{[VOID]} padding.
  During inference, \texttt{[VOID]} generation is banned. We introduce
  \textbf{\texttt{[EOS]}Termination}, which stops decoding once no masks remain before the
  leftmost visible \texttt{[EOS]}.
  This prevents \texttt{[EOS]} overflow and avoids redundant generation after the
  effective response.
  Additionally, we propose an optional inference-time 
  \textbf{VoidExpansion} strategy, which uses the \texttt{[VOID]}
  length signal to expand insufficient canvases before denoising.

 Our contributions are:
  \begin{enumerate}
      \item We systematically analyze the mechanism behind \texttt{[EOS]} overflow and
      show that it stems from the dual use of \texttt{[EOS]} for termination and
      padding during instruction tuning.
  \item We propose VoidPadding, a training--inference co-design that decouples
        the two roles by using \texttt{[VOID]} for padding and reserving
        \texttt{[EOS]} for termination, enabling \texttt{[EOS]}Termination for
        early stopping and VoidExpansion for adaptive canvas expansion.
  \item Experiments show that VoidPadding prevents \texttt{[EOS]} overflow,
        improves generation quality and decoding efficiency, and remains robust
        across block sizes and canvas lengths.
  \end{enumerate}

\section{Preliminary}
\subsection{Masked Diffusion Modeling}
\label{subsec:masked-diffusion-modeling}
  MDLMs generate by denoising discrete
  sequences~\citep{shi2024simplified,sahoo2024simple}.
  Let \(\mathcal{V}=\{1,\ldots,V\}\) be the token vocabulary, \(\mathsf{M}\notin\mathcal{V}\)
  the mask token, and \(\mathbf{z}_t\in(\mathcal{V}\cup\{\mathsf{M}\})^L\) the sequence state
  at diffusion step \(t\in\{0,\ldots,T\}\), where \(T\) is the total number of denoising steps.
  Here, \(t=0\) denotes clean data and \(t=T\) the fully masked sequence.
  The masked and visible positions are denoted by \(\mathcal{M}_t=\{i:z_t^i=\mathsf{M}\}\) and
  \(\bar{\mathcal{M}}_t=\{1,\ldots,L\}\setminus\mathcal{M}_t\), respectively.

The forward process gradually corrupts a clean sequence \(\mathbf{z}_0\sim p_{\mathrm{data}}\) by independently masking each position over \(T\) steps.
At each step, tokens are progressively replaced by \(\mathsf{M}\) according to a fixed schedule, so that \(\mathbf{z}_T\) is fully masked.
During training, a timestep \(t\) is sampled and the model predicts the original token at each masked position, yielding the denoising cross-entropy objective
\(
    \mathcal{L}(\theta)
    =
    \mathbb{E}_{\mathbf{z}_0,t,\mathbf{z}_t}
    \left[
    \sum_{i\in\mathcal{M}_t}
    -\log p_\theta(z_0^i\mid\mathbf{z}_t)
    \right].
\)
  This objective trains the denoiser on various masking patterns, enabling
  \(p_\theta\) to predict masked tokens from different visible contexts~\citep{kim2025train}.

MDLM decoding procedures include block diffusion~\citep{arriola2025block} and pure
  diffusion~\citep{ye2025dream}.
  Both of them start from a fully masked canvas of length \(L\) with block size \(B\),
  where pure diffusion corresponds to \(B=L\).
  At each step, the model uses bidirectional attention to condition on the whole sequence
  and predicts \(p_\theta(\cdot\mid\mathbf{z}_t)\) for all \(i\in\mathcal{M}_t\) in parallel,
  so one denoising step corresponds to one model forward pass.
 Decoding then proceeds in two steps within the current block.
  First, it selects commitment positions, commonly using
  confidence~\cite{cai2026confidence} or negative entropy scores~\citep{ben2026accelerated}.
  Second, it selects tokens for these positions, typically by argmax under
  \(p_\theta\).
  Decoding moves to the next block once the current block is fully committed and terminates
  when all masks have been filled.

\subsection{Instruction-Tuning Data Construction}
\label{subsec:eos-padding-instruction-tuning}
Instruction tuning (IT) trains pretrained language models on prompt--response pairs,
  improving instruction following and zero-shot generalization~\cite{chung2024scaling}.
  For efficient batching, prompt--response sequences are terminated by
  \texttt{[EOS]} and padded to a common length.
 MDLM IT often follows the ARLM convention of using \texttt{[EOS]} as the padding token.
  Let \(B=\{(p_k,r_k)\}_{k=1}^{b}\) be a mini-batch of size \(b\), where \(p_k\) is the prompt
  of length \(P_k\) and \(r_k\) is the raw response of length \(\ell_k\).
  Over the IT data distribution, let \(\ell\) denote the raw response length before the
  terminal \texttt{[EOS]}, and define its CDF as \(F(m)=\Pr(\ell\leq m)\).
  After appending one semantic terminator \texttt{[EOS]}, each full sequence is padded with
  \texttt{[EOS]} to the maximum full-sequence length in the mini-batch, \(S_B=\max_{1\leq k\leq
  b}(P_k+\ell_k+1)\).
  Thus, example \(k\) has padded response \(y_k=r_k\oplus \texttt{[EOS]}\oplus \texttt{[EOS]}
  ^{L_{B,k}-\ell_k-1}\) of length \(L_{B,k}=S_B-P_k\).
  During IT, both noise and loss are applied only to \(y_k\).

\section{Analysis of \texttt{[EOS]} Overflow}
\label{sec:eos-analysis}
\begin{figure}[t]
\centering
\definecolor{panelGrayBorder}{HTML}{D8DDE3}
\definecolor{panelGrayFill}{HTML}{F6F7F8}
\definecolor{janetBlue}{HTML}{5AA9E6}
\definecolor{eosTokenRed}{HTML}{B3342A}
\scriptsize
\setlength{\fboxsep}{1pt}
\begin{subfigure}[t]{0.94\linewidth}
\centering
\fcolorbox{panelGrayBorder}{panelGrayFill}{
\begin{minipage}{0.985\linewidth}
{\fontsize{6.3pt}{6.7pt}\selectfont\textbf{(a) Original model} \hfill \texttt{[EOS]} overflow}\\[-3pt]
{\fontsize{4.6pt}{4.8pt}\selectfont\ttfamily \textcolor{janetBlue}{Janet} \textcolor{eosTokenRed}{<eos><eos><eos><eos><eos><eos><eos><eos>} ...}
\end{minipage}}
\\[-1pt]
\includegraphics[width=\linewidth]{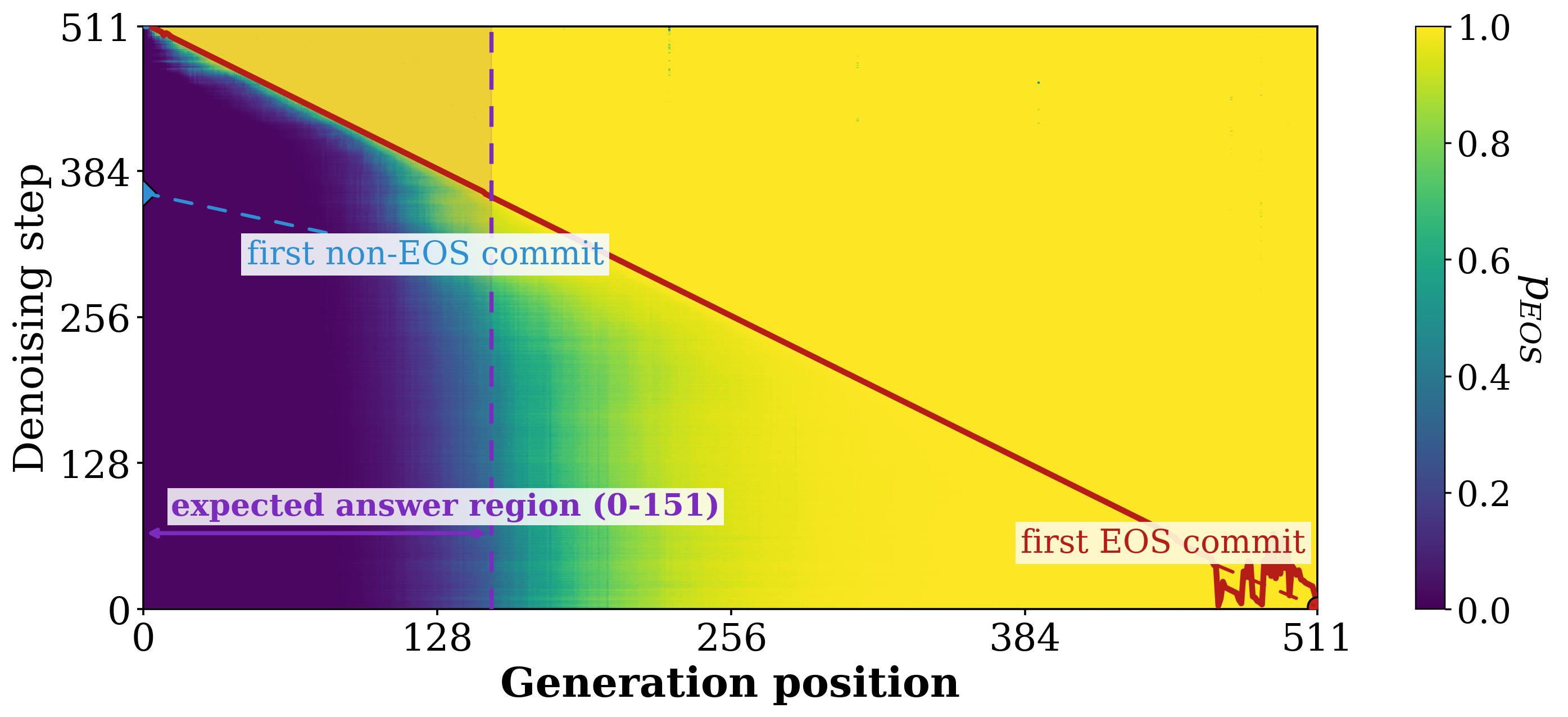}
\end{subfigure}

\vspace{2.5pt}
\begin{subfigure}[t]{0.94\linewidth}
\centering
\fcolorbox{panelGrayBorder}{panelGrayFill}{
\begin{minipage}{0.985\linewidth}
{\fontsize{6.3pt}{6.7pt}\selectfont\textbf{(b) Masking attention to committed \texttt{[EOS]}} \hfill Correct generation}\\[-3pt]
{\fontsize{4.6pt}{4.8pt}\selectfont\ttfamily \textcolor{janetBlue}{Janet's ducks lay 16 eggs per day. She eats 3 eggs for breakfast every morning, so she has}}\\[-2.5pt]
{\fontsize{4.6pt}{4.8pt}\selectfont\ttfamily \textcolor{janetBlue}{16 - 3 = 13 eggs left. She bakes muffins with 4 eggs, so she has 13 - 4 = 9 eggs left. She}}\\[-2.5pt]
{\fontsize{4.6pt}{4.8pt}\selectfont\ttfamily \textcolor{janetBlue}{ sells 9 eggs for \$2 each. Therefore, she makes 9 * \$2 = \$18. \#\#\#\# 18} \textcolor{eosTokenRed}{<eos><eos><eos> }...}
\end{minipage}}
\\[-1pt]
\includegraphics[width=\linewidth]{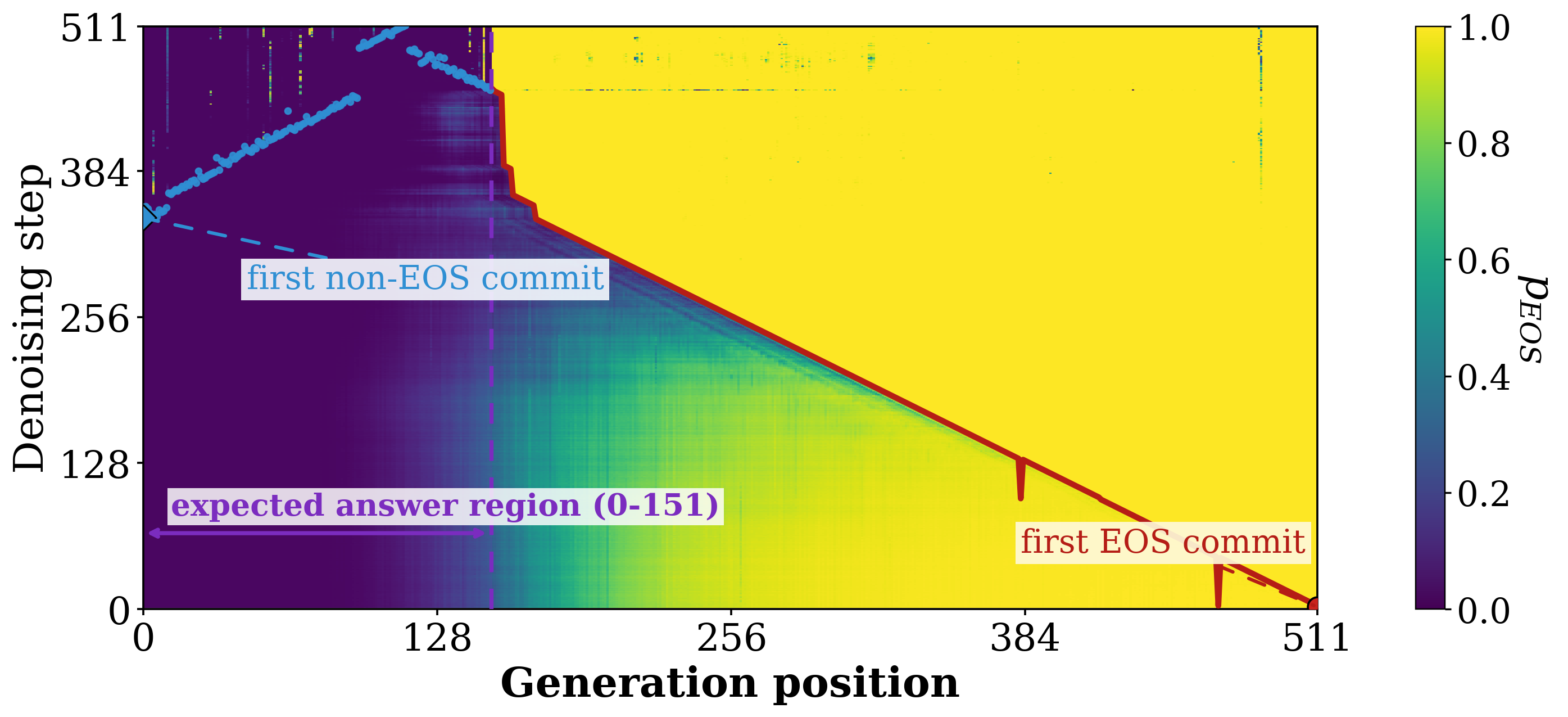}
\end{subfigure}
\caption{Attention intervention on the \(B=L=512\) \texttt{[EOS]} overflow trajectory. Both cases use the same prompt. Masking
  attention to committed \texttt{[EOS]} tokens avoids \texttt{[EOS]} overflow and recovers the correct answer.}
\label{fig:eos-attention-intervention}
\end{figure}

  \texttt{[EOS]} overflow is a common issue of MDLM inference under large block sizes~\citep{kim2025rainbow}.
  In a typical case, the model produces a very short answer followed by repeated \texttt{[EOS]} tokens, as shown in Figure~\ref{fig:eos-attention-intervention}(a).
  However, \texttt{[EOS]} overflow is not observed in ARLMs, even though  both ARLMs and MDLMs use \texttt{[EOS]} as padding tokens during IT.
  This  motivates us to ask: 
 Why is \texttt{[EOS]} padding safe in ARLMs but fragile in MDLMs?
\begin{observationbox}
\textbf{
  Observation 1: \texttt{[EOS]} overflow is not caused by limited large-block generation capacity.}
\end{observationbox}
We first test whether \texttt{[EOS]} overflow is caused by insufficient large-block generation capacity.
Specifically, we ban \texttt{[EOS]} commitment in LLaDA-8B-Instruct by setting the \texttt{[EOS]} logit to \(-\infty\) throughout decoding, and then evaluate on GSM8K~\cite{cobbe2021training} with \(L=B=512\).
Although this removes the semantic stopping signal and often causes repeated reasoning traces and answers, it raises the accuracy of the first extracted answer from 28.43\% to 70.66\%.
Thus, the model is not inherently unable to generate large blocks.

\begin{figure}[t]
\centering
\centering
\includegraphics[width=\linewidth]{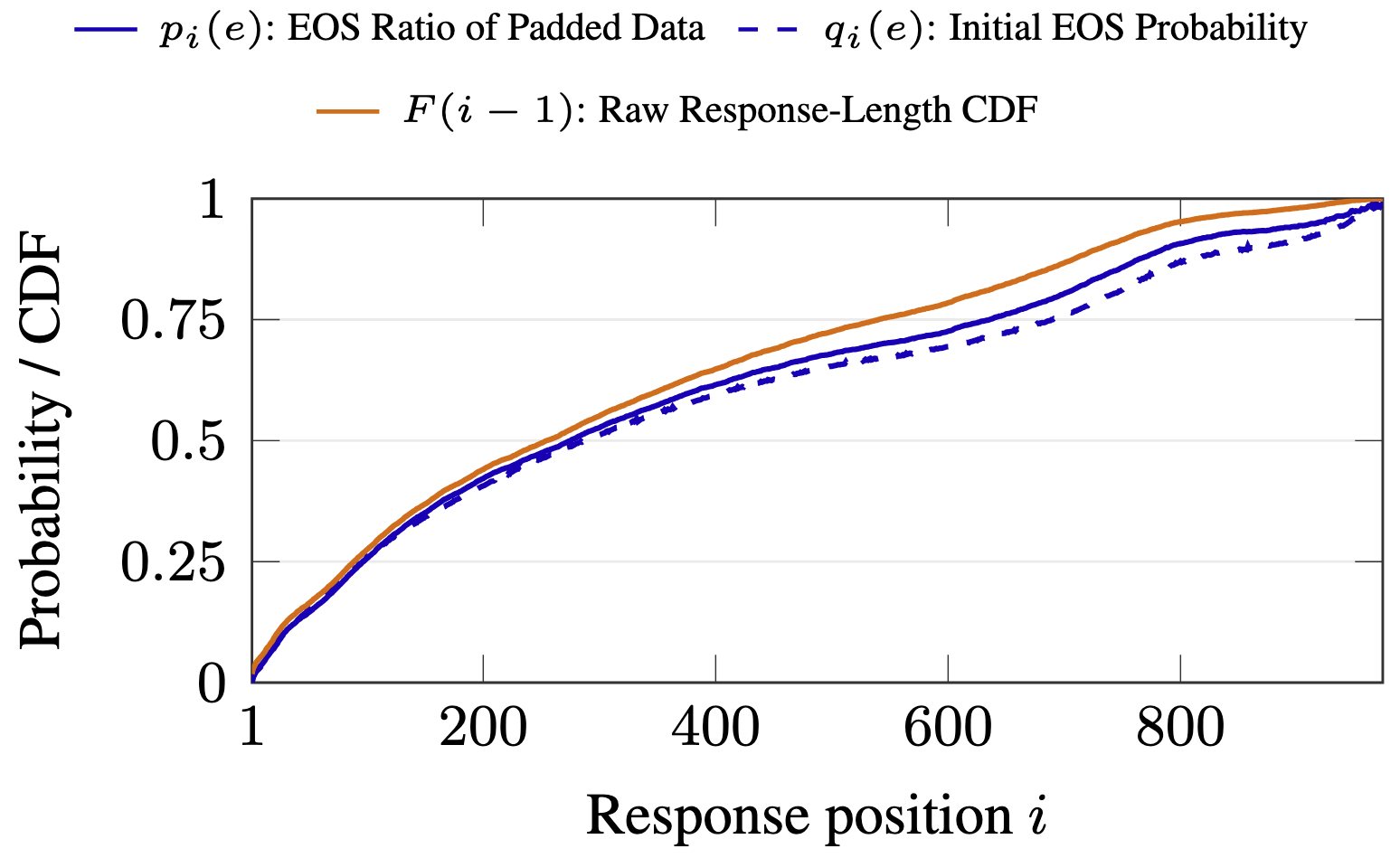}
\caption{\texttt{[EOS]} as a learned length signal. Initial \texttt{[EOS]} confidence aligns with the \texttt{[EOS]} label ratio of padded data and the raw response-length CDF.}
\label{fig:training-aligned-eos-prior-probe}
\end{figure}
\begin{observationbox}
  \phantomsection\label{obs:eos-hidden-length-signal}
 \textbf{Observation 2: Training with \texttt{[EOS]} padding teaches \texttt{[EOS]} to act as a length signal.}
\end{observationbox}
  Next, we examine whether \texttt{[EOS]} padding provides an additional signal in MDLM
  training beyond semantic termination.
  Specifically, we train an \texttt{[EOS]}-padding diagnostic model on LLaDA-8B-Base, with
  training details provided in Appendix~\ref{app:training-configuration}.
  Following Section~\ref{subsec:eos-padding-instruction-tuning}, for each mini-batch
  \(B=\{(p_k,r_k)\}_{k=1}^{b}\), we construct the padded response
  \(
      y_k
      =
      r_k
      \oplus \texttt{[EOS]}
      \oplus \texttt{[EOS]}^{L_{B,k}-\ell_k-1},
  \)
  where \(L_{B,k}=S_B-P_k\) and \(S_B=\max_{1\leq k\leq b}(P_k+\ell_k+1)\).
  We define response-side positions using a 1-indexed coordinate \(i\in\{1,\dots,1024\}\).
  For the padded data distribution, we record the raw response-length CDF \(F(i-1)=\Pr(\ell\leq
  i-1)\) defined in Section~\ref{subsec:eos-padding-instruction-tuning}, along with the
  \texttt{[EOS]} token ratio \(p_i(e)\) at response-side position \(i\).
  For the model distribution, we use the same batch-padded examples, mask the entire padded
  response \(y_k\) while keeping the prompt \(p_k\) visible, and report the initial
  \texttt{[EOS]} probability \(q_i(e)\) from a single forward pass through the diagnostic
  model.
  As shown in Figure~\ref{fig:training-aligned-eos-prior-probe}, the three curves closely
  align, \(F(i-1)\approx p_i(e)\approx q_i(e)\), indicating that \texttt{[EOS]} padding teaches
  the model to associate \texttt{[EOS]} with desirable response length.

\begin{observationbox}
\textbf{Observation 3: Large--block inference converts the  \texttt{[EOS]} length signal into overflow.}
\end{observationbox}
We analyze a large-block decoding trajectory of LLaDA-8B-Instruct with
\(L=B=512\).  Figure~\ref{fig:eos-attention-intervention}(a) shows a
two-stage \texttt{[EOS]} overflow pattern.  In Stage 1, on the initial fully
 masked canvas, tail positions have very high \texttt{[EOS]} probability and
\texttt{[EOS]} is committed in the initial steps.  In Stage 2, on a
canvas with a committed \texttt{[EOS]} suffix, \texttt{[EOS]} commitments move
leftward: positions before the suffix also take \texttt{[EOS]} as their
committed token, producing overflow.
We then formalize this pattern under the idealized version of
Observation~\hyperref[obs:eos-hidden-length-signal]{2}.
Let \(i\) index response-side positions and let \(e\) denote \texttt{[EOS]}.
For a training-time EOS-padded target, let \(\bar{Y}_i\) be the token at
position \(i\), and write \(p_i(e)=\Pr(\bar{Y}_i=e)\).  During inference, let
\(q_i\) be the model distribution at masked canvas position \(i\), let
\(\hat{y}_i=\arg\max_v q_i(v)\) be its proposed token, and let \(Y_i\) be the
token committed at canvas position \(i\).

For Stage 1, on a fully masked response canvas, we idealize
Observation~\hyperref[obs:eos-hidden-length-signal]{2} as
\[
    q_i(e)=p_i(e)=F(i-1).
\]  Since standard MDLM decoding uses argmax token
proposals (Section~\ref{subsec:masked-diffusion-modeling}), \(q_i(e)>1/2\)
implies \(\hat{y}_i=e\).
  Assume the canvas is long enough that \(F(L-1)>1/2\). On the initial fully 
  masked response canvas, we call the suffix of positions with \(q_i(e)>1/2\) the
  high-confidence \texttt{[EOS]} tail, and define its beginning as
  \[
      r=\min\{i\in\{1,\dots,L\}: F(i-1)>1/2\}.
  \]

For Stage 2, let \(S_e=\{Y_r=\cdots=Y_L=e\}\) denote the committed
\texttt{[EOS]} suffix event during inference, and let
\(\bar{S}_e=\{\bar{Y}_r=\cdots=\bar{Y}_L=e\}\) denote the corresponding
training-time \texttt{[EOS]}  padding suffix event.  
  We assume that after \(S_e\) occurs, the model's \texttt{[EOS]} probability for
  earlier positions matches the corresponding training-time \texttt{[EOS]}-padded
  conditional probability: for any \(i<r\),
  \[
      q_i(e\mid S_e)
      =
      p_i(e\mid \bar{S}_e)
      =
      \Pr(\bar{Y}_i=e\mid \bar{S}_e).
  \]
\begin{theorem}[Stage 1: initial \texttt{[EOS]} commitments]
\label{thm:tail-eos-commit}
 If the first block reaches the high-confidence \texttt{[EOS]} tail, \(r\leq B\),
  then every selected position \(j\in\{r,\dots,B\}\) is committed as
  \texttt{[EOS]}, i.e., \(Y_j=e\).
\end{theorem}

\begin{theorem}[Stage 2: leftward spread of \texttt{[EOS]} commitments]
\label{thm:leftward-eos-spread}
If \(S_e\) has occurred, then for any \(i<r\) satisfying
\(2F(i-1)>F(r-1)\), selecting position \(i\) in a later denoising step yields
\(Y_i=e\).
\end{theorem}
Proofs of the theorems are given in
Appendix~\ref{sec:proof}.

We next perform an intervention motivated by
Theorem~\ref{thm:leftward-eos-spread}: after a position is committed as
\texttt{[EOS]}, we set attention scores to that position to \(-\infty\), as
shown in Figure~\ref{fig:eos-attention-intervention}(b).
Initial \texttt{[EOS]} commitments remain, but the model avoids
\texttt{[EOS]} overflow and generates the correct answer.  This shows that Stage 2
is the dominant driver of \texttt{[EOS]} overflow.

  These observations reveal that \texttt{[EOS]} overflow stems from a
  training--inference role mismatch. In MDLM IT, bidirectional attention exposes
  padding \texttt{[EOS]} tokens as denoising targets, so \texttt{[EOS]} is trained
  as both a semantic terminator and a padding token. During inference, however,
  \texttt{[EOS]} is expected to serve only as a semantic terminator. In ARLM IT, causal masking hides
  padding \texttt{[EOS]} tokens from earlier positions, so \texttt{[EOS]} remains
  a semantic terminator and overflow does not arise.

\section{Method}

 Motivated by this training--inference role mismatch, we propose
  \textbf{VoidPadding}, which introduces a new \texttt{[VOID]} token as the
  padding token, so that \texttt{[EOS]} serves only as the semantic terminator.
  We further design \textbf{VoidExpansion} for inference, which uses the
  \texttt{[VOID]} length signal learned during IT to expand an insufficient
canvas before decoding.
  Together, this training--inference co-design decouples padding from semantic
  termination while preserving the hidden length signal carried by
  \texttt{[VOID]} tokens.
  Figure~\ref{fig:void-padding-overview} provides an overview of the method, with
  details given in the following subsections.
  \subsection{VoidPadding}
  \label{subsec:void-padding}
  \noindent\textbf{Instruction-Tuning Design.} VoidPadding follows the IT data construction
  described in Section~\ref{subsec:eos-padding-instruction-tuning}, except that
  padding \texttt{[EOS]} tokens are replaced by \texttt{[VOID]} tokens. Thus, for
  \(B=\{(p_k,r_k)\}_{k=1}^{b}\) and \(L_{B,k}=S_B-P_k\), the padded response is
  \[
      y_k^{\mathrm{VOID}}
      =
      r_k
      \oplus \texttt{[EOS]}
      \oplus \texttt{[VOID]}^{L_{B,k}-\ell_k-1}.
  \]
  We then follow the standard MDLM training procedure in
  Section~\ref{subsec:masked-diffusion-modeling}, with noise injection and loss
  computation performed on \(y_k^{\mathrm{VOID}}\).
  Notably, we apply standard MDLM masking for non-\texttt{[EOS]} positions.
  The single \texttt{[EOS]} token in each example is always masked, forcing the
  model to predict it as a stable semantic terminator.

 \noindent\textbf{Inference Design.}
  During inference, \texttt{[VOID]} remains a hidden padding signal and is never
  generated: setting its logit to \(-\infty\) before token selection prevents the
  high-confidence \texttt{[VOID]} tail from becoming visible output.
  We also propose \textbf{\texttt{[EOS]}Termination}, which stops decoding once
  all positions before the leftmost visible \texttt{[EOS]} are unmasked and then
  discards the rest of the canvas.
  In contrast, \textbf{VanillaStopping}, the standard MDLM stopping rule, stops
  only after all masked positions are decoded
  (Section~\ref{subsec:masked-diffusion-modeling}), causing redundant decoding
  after the semantic endpoint.

\subsection{VoidExpansion}
\label{subsec:void-expansion}
  As discussed in Section~\ref{subsec:masked-diffusion-modeling}, MDLM decoding
  starts from a pre-defined response canvas with fixed length \(L\) and block size
  \(B\). When this canvas is too short, generation quality drops sharply
  (Appendix Figure~\ref{fig:instruct-genlen-sweep}).
  Motivated by this short-canvas degradation, VoidExpansion uses the learned
  \texttt{[VOID]} length signal to expand the canvas to a sufficient, block-aligned length
  \(\hat L=nB\) before decoding.

Similar to how IT with \texttt{[EOS]} padding induces a length signal in
  \texttt{[EOS]} (Observation~\hyperref[obs:eos-hidden-length-signal]{2}), IT with
  \texttt{[VOID]} padding induces a \texttt{[VOID]} length signal.
  Specifically, sufficient canvases have redundant tails that prefer
  \texttt{[VOID]}, whereas insufficient canvases retain tail positions that favor
  non-\texttt{[VOID]} tokens.
  We therefore quantify this \texttt{[VOID]} length signal as follows.
  Let \(p\) denote the prompt tokens, \(\mathsf{M}\) the mask token, and \(w\) the
  tail-window size.
  At canvas length \(\ell\), we run one forward pass on
  \(p\oplus\mathsf{M}^{\ell}\) and define the tail window
  \(\mathcal{T}(\ell)\) as the last \(\min(w,\ell)\) response-canvas positions,
  indexed by \(i\):
\[
\begin{aligned}
  p_i^{\mathrm{void}}
  &= p_\theta(x_i=\texttt{[VOID]} \mid p\oplus\mathsf{M}^{\ell}),\\
  p_i^{\mathrm{nonvoid}}
  &= \max_{a\in\mathcal{V}\setminus\{\texttt{[VOID]}\}}
     p_\theta(x_i=a \mid p\oplus\mathsf{M}^{\ell}),\\
  s_{\mathrm{nonvoid}}(\ell)
  &= |\mathcal{T}(\ell)|^{-1}\sum_{i\in\mathcal{T}(\ell)}p_i^{\mathrm{nonvoid}},\\
  s_{\mathrm{gap}}(\ell)
  &= |\mathcal{T}(\ell)|^{-1}\sum_{i\in\mathcal{T}(\ell)}
     \left(p_i^{\mathrm{void}}-p_i^{\mathrm{nonvoid}}\right).
\end{aligned}
\]
  Here \(s_{\mathrm{nonvoid}}\) measures tail preference for non-\texttt{[VOID]}
  tokens, while \(s_{\mathrm{gap}}\) measures tail preference for \texttt{[VOID]}
  over the strongest non-\texttt{[VOID]} alternative at each position.
  Thus, high \(s_{\mathrm{nonvoid}}\) or low \(s_{\mathrm{gap}}\) suggests that
  the current canvas is insufficient.

 VoidExpansion uses this \texttt{[VOID]} length signal to expand the insufficient
  canvas, as shown in Algorithm~\ref{alg:void-expansion}.
  It defines a canvas as sufficient when
  \(
    s_{\mathrm{nonvoid}}(L) \le \tau_{\mathrm{nonvoid}}
    \: \text{and} \:
    s_{\mathrm{gap}}(L) \ge \tau_{\mathrm{gap}}.
  \)
  Starting from the block-aligned initial canvas length \(L=\lceil L_0/B\rceil B\),
  VoidExpansion checks this criterion at the current \(L\). If the criterion is not
  met, it expands the canvas by one block and repeats until the criterion is met or
  the maximum length restriction \(L_{\mathrm{cap}}=\lceil L_{\max}/B\rceil B\) is reached.
\begin{algorithm}[t]
\small
\caption{VoidExpansion Before Decoding}
\label{alg:void-expansion}
\begin{algorithmic}[1]
\Require Prompt tokens \(p\), model \(f_\theta\), initial canvas length \(L_0\), maximum length \(L_{\max}\), block size \(B\), tail window \(w\), thresholds \(\tau_{\mathrm{nonvoid}}\) and \(\tau_{\mathrm{gap}}\)
\Ensure Selected generation length \(\hat L\)
\State \(L \gets \lceil L_0/B\rceil B\), \(L_{\mathrm{cap}}\gets\lceil L_{\max}/B\rceil B\)
\While{\(L \le L_{\mathrm{cap}}\)}
    \State Construct \(x = p \oplus \mathsf{M}^L\)
    \State \(z \gets f_\theta(x)_{\mathrm{resp}}\) \Comment{logits on response positions}
    \State \(\mathcal{T}(L) \gets\) last \(\min(w,L)\) positions of \(z\)
    \State Compute \(s_{\mathrm{nonvoid}}(L)\) and \(s_{\mathrm{gap}}(L)\) on \(\mathcal{T}(L)\)
    \If{\(s_{\mathrm{nonvoid}}(L) \le \tau_{\mathrm{nonvoid}}\) \textbf{and} \(s_{\mathrm{gap}}(L) \ge \tau_{\mathrm{gap}}\)}
        \State \Return \(L\)
    \EndIf
    \State \(L \gets L + B\) \Comment{append one block before generation}
\EndWhile
\State \Return \(L_{\mathrm{cap}}\)
\end{algorithmic}
\end{algorithm}

\section{Experiments}
\label{sec:experiments}

\noindent\textbf{Setup.}
Our main experiments use two instruction-tuned MDLMs: LLaDA-8B-Instruct (hereafter LLaDA)~\citep{nie2026large} and Dream-7B-Instruct (hereafter Dream)~\citep{ye2025dream}. Quality metrics are accuracy on
  GSM8K~\citep{cobbe2021training} and MATH500~\citep{hendrycks2021measuring},
  and pass@1 on HumanEval~\citep{chen2021evaluating} and
  MBPP~\citep{austin2021program}.
  The main efficiency metric is the average number of forward
  passes per example (hereafter NFE).
 In all tables, bold marks the best
result and underline marks the second best within each comparison.
  For Dream, we use a decoding wrapper to enable the same block-wise decoding
  protocol as LLaDA; details are provided in
  Appendix~\ref{app:dream-wrapper}. We compare against the original models and
  RainbowPadding~\citep{kim2025rainbow}, 
  which aims to avoid 
  \texttt{[EOS]} overflow by cyclically using 7 distinct padding
  tokens.
 On both LLaDA and Dream, RainbowPadding and VoidPadding use
  LoRA SFT~\citep{hu2022lora} on the same 0.5M-example data and identical
  training configurations, except that VoidPadding uses 67\% of
  RainbowPadding's training steps. Detailed training configuration is provided in
  Appendix~\ref{app:training-configuration}.

\subsection{Fixed-Length Block-Size Robustness}
\label{subsec:fixed-length-block-size-robustness}

  We first evaluate fixed-length robustness with generation length \(L=512\) and a maximum
  denoising-step budget \(T=512\), while varying the block size \(B\).
  Table~\ref{tab:blocksize-robustness-main} reports
\(B\in\{64,128,512\}\).
Complete block-size results are in
Appendix~\ref{app:llada8b-instruct-blocksize-sweeps}. 
We compare against the original LLaDA and Dream models, as well as RainbowPadding applied to these models.

\begin{table*}[t]
\centering
\scriptsize
\begin{subtable}[t]{0.49\textwidth}
\centering
\setlength{\tabcolsep}{2.5pt}
\resizebox{\linewidth}{!}{
\begin{tabular}{llrrrr}
\toprule
\multirow{2}{*}{Task} &
\multirow{2}{*}{Method} &
\multicolumn{3}{c}{Block size \(B\)} &
\multirow{2}{*}{Mean} \\
\cmidrule(lr){3-5}
& & 64 & 128 & 512 & \\
\midrule
\multirow{3}{*}{GSM8K}
& Original& \underline{76.80} & 70.96 & 28.43 & 58.73 \\
& RainbowPadding & 74.53 & \underline{73.39} & \underline{66.57} & \underline{71.50} \\
& VoidPadding & \textbf{77.79} & \textbf{78.39} & \textbf{78.62} & \textbf{78.27} \\
\addlinespace[2pt]
\multirow{3}{*}{MATH500}
& Original& \underline{39.60} & \underline{37.60} & 16.60 & 31.27 \\
& RainbowPadding & 33.20 & 34.00 & \textbf{27.80} & \underline{31.67} \\
& VoidPadding & \textbf{40.60} & \textbf{41.60} & \underline{27.60} & \textbf{36.60} \\
\addlinespace[2pt]
\multirow{3}{*}{HEval}
& Original& \underline{42.68} & \textbf{43.90} & \textbf{44.51} & \textbf{43.70} \\
& RainbowPadding & \textbf{43.90} & \underline{43.29} & 35.98 & 41.06 \\
& VoidPadding & 41.46 & 42.07 & \underline{41.46} & \underline{41.66} \\
\addlinespace[2pt]
\multirow{3}{*}{MBPP}
& Original& 41.48 & 39.12 & 37.68 & 39.43 \\
& RainbowPadding & \underline{42.71} & \underline{43.02} & \underline{42.51} & \underline{42.75} \\
& VoidPadding & \textbf{44.66} & \textbf{44.87} & \textbf{44.97} & \textbf{44.83} \\
\addlinespace[2pt]
\multirow{3}{*}{Mean over tasks}
& Original& \underline{50.14} & 47.90 & 31.81 & 43.28 \\
& RainbowPadding & 48.59 & \underline{48.43} & \underline{43.22} & \underline{46.75} \\
& VoidPadding & \textbf{51.13} & \textbf{51.73} & \textbf{48.16} & \textbf{50.34} \\
\bottomrule
\end{tabular}
}
\caption{LLaDA}
\label{tab:blocksize-robustness-instruct}
\end{subtable}
\hfill
\begin{subtable}[t]{0.49\textwidth}
\centering
\setlength{\tabcolsep}{2.5pt}
\resizebox{\linewidth}{!}{
\begin{tabular}{llrrrr}
\toprule
\multirow{2}{*}{Task} &
\multirow{2}{*}{Method} &
\multicolumn{3}{c}{Block length \(B\)} &
\multirow{2}{*}{Mean} \\
\cmidrule(lr){3-5}
& & 64 & 128 & 512 & \\
\midrule
\multirow{3}{*}{GSM8K}
& Original & 73.39 & 33.89 & 46.70 & 51.33 \\
& RainbowPadding & \underline{79.15} & \underline{77.63} & \underline{52.16} & \underline{69.65} \\
& VoidPadding & \textbf{80.44} & \textbf{80.89} & \textbf{78.09} & \textbf{79.81} \\
\addlinespace[2pt]
\multirow{3}{*}{MATH500}
& Original & \underline{43.20} & \underline{41.20} & 26.00 & \underline{36.80} \\
& RainbowPadding & 33.60 & 33.40 & \underline{31.40} & 32.80 \\
& VoidPadding & \textbf{44.00} & \textbf{44.20} & \textbf{45.20} & \textbf{44.47} \\
\addlinespace[2pt]
\multirow{3}{*}{HEval}
& Original & \textbf{62.80} & \textbf{60.37} & 18.90 & 47.36 \\
& RainbowPadding & \textbf{62.80} & \textbf{60.37} & \underline{50.00} & \underline{57.72} \\
& VoidPadding & \underline{59.76} & \underline{59.15} & \textbf{60.37} & \textbf{59.76} \\
\addlinespace[2pt]
\multirow{3}{*}{MBPP}
& Original & 41.00 & 31.00 & 30.00 & 34.00 \\
& RainbowPadding & \underline{56.00} & \underline{54.80} & \underline{47.80} & \underline{52.87} \\
& VoidPadding & \textbf{57.60} & \textbf{58.00} & \textbf{54.80} & \textbf{56.80} \\
\addlinespace[2pt]
\multirow{3}{*}{Mean over tasks}
& Original & 55.10 & 41.62 & 30.40 & 42.37 \\
& RainbowPadding & \underline{57.89} & \underline{56.55} & \underline{45.34} & \underline{53.26} \\
& VoidPadding & \textbf{60.45} & \textbf{60.56} & \textbf{59.62} & \textbf{60.21} \\
\bottomrule
\end{tabular}
}
\caption{Dream}
\end{subtable}

\caption{Fixed-length generation robustness experiments on LLaDA and Dream with \(B\in\{64,128,512\}\).}
\label{tab:blocksize-robustness-main}
\end{table*}

\noindent\textbf{Accuracy Robustness.}
 Table~\ref{tab:blocksize-robustness-main} shows that VoidPadding improves fixed-length robustness across both
 models.
  On LLaDA, \texttt{[VOID]} substantially mitigates large-block \texttt{[EOS]} overflow: the original model drops sharply from
  76.80 at \(B=64\) to 28.43 at \(B=512\) on GSM8K, while \texttt{[VOID]} remains around 78.
It gives the strongest mean on three of
four tasks and remains competitive on HumanEval. Dream shows the same trend: the original model degrades as \(B\) grows, while \texttt{[VOID]} remains stable and attains the best mean score across tasks.
  Across all block sizes, \texttt{[VOID]} attains the highest four-task mean.
On LLaDA, the block-size-averaged macro mean over four tasks improves by
\( +7.06\) points over the original model and \(+3.59\) over Rainbow.
On Dream, the corresponding block-size-averaged macro means are \(+17.84\) and \(+6.95\), respectively.
\begin{table}[t]
\centering
\scriptsize
\setlength{\tabcolsep}{4pt}
\resizebox{\columnwidth}{!}{
\begin{tabular}{lrrr}
\toprule
Benchmark & Original/Rainbow & VoidPadding & NFE Reduction \\
\midrule
GSM8K & 512.0 & 157.3 & 69.3\% \\
MATH500 & 512.0 & 371.0 & 27.5\% \\
HumanEval & 512.0 & 300.7 & 41.3\% \\
MBPP & 512.0 & 78.7 & 84.6\% \\
\bottomrule
\end{tabular}
}
\caption{NFE for Dream variants in Table~\ref{tab:blocksize-robustness-main}, averaged over \(B\in\{64,128,512\}\).}
\label{tab:fixed-length-efficiency-main}
\end{table}

 \noindent\textbf{\texttt{[EOS]}Termination Efficiency.} 
  Table~\ref{tab:fixed-length-efficiency-main} reports the average NFE of
  Dream variants in Table~\ref{tab:blocksize-robustness-main} over
  \(B\in\{64,128,512\}\).
  The original model and RainbowPadding use VanillaStopping, which fills the entire
  512-token canvas. Since each denoising step requires one model forward pass,
  this setting requires 512 NFE under \(T=512\).
  VoidPadding instead uses \texttt{[EOS]}Termination, which reduces NFE by 69.3\% on GSM8K, 27.5\% on MATH500, 41.3\% on
  HumanEval, and 84.6\% on MBPP, averaged over \(B\in\{64,128,512\}\).
  Appendix~\ref{app:fixed-length-efficiency} reports per-block NFE results for
  both LLaDA and Dream.

  \subsection{Effectiveness of VoidExpansion}

  We next evaluate VoidExpansion on a short initial canvas and compare it
  against two inference-only variable-length generation methods:
  \(\rho\)-\texttt{[EOS]}~\citep{yang2026rho} and
  \textsc{Daedal}~\citep{li2025beyond}. Both methods adjust the canvas using the
  \texttt{[EOS]} signal of the original model.
  In the main short-canvas experiment, all methods start from an initial length
  \(L_0=64\). VoidExpansion uses a maximum length
  \(L_{\max}=512\), block size \(B=32\), and tail-window size \(w=16\).
  On LLaDA, fixed-\(L_0\) decoding baselines are evaluated on both
  the original model and the \texttt{[VOID]}-padding checkpoint.
  We apply \(\rho\)-\texttt{[EOS]} and \textsc{Daedal} to the original model, and
  VoidExpansion to the \texttt{[VOID]}-padding checkpoint, comparing
  our \texttt{[VOID]}-based training--inference co-design with
  \texttt{[EOS]}-based inference-only length selection.
  For fixed-\(L_0\) decoding and VoidExpansion, the maximum
  denoising-step budget is \(T=L_0\) and \(T=\hat L\), respectively, where
  \(\hat L\) is the canvas length after VoidExpansion. For \(\rho\)-\texttt{[EOS]} and
  \textsc{Daedal}, we follow their adaptive decoding procedures. In all cases,
  NFE counts only decoding forward passes, while wall-clock time includes both
  canvas selection and decoding. Since the official
  implementations of \(\rho\)-\texttt{[EOS]} and \textsc{Daedal} do not support
  Dream, Dream is compared only against fixed-\(L_0\) decoding.

\begin{table*}[t]
\centering
\scriptsize
\setlength{\tabcolsep}{3pt}
\definecolor{wallSpeedBlue}{HTML}{1F5AA6}
\newcommand{\wallspd}[1]{\textcolor{wallSpeedBlue}{#1\(\times\)}}

\begin{subtable}[t]{\textwidth}
\centering
\begin{tabular*}{0.95\textwidth}{@{\extracolsep{\fill}}lccrrrrr@{}}
\toprule
Method & Fine-tuned & Expansion Signal & GSM8K & MATH500 & HEval & MBPP & Mean \\
\midrule
LLaDA fixed-64 & No & None & 60.20 & 26.80 & 29.27 & 41.27 & 39.38 \\
VOID fixed-64 & Yes & None & 52.84 & 26.20 & 36.59 & 38.81 & 38.61 \\
\(\rho\)-\texttt{[EOS]} & No & \texttt{[EOS]} & 73.39 & 27.00 & 40.24 & 41.38 & 45.50 \\
\textsc{Daedal} & No & \texttt{[EOS]} & \textbf{79.38} & \underline{37.40} & \textbf{44.51} & \underline{41.79} & \underline{50.77} \\
VoidExpansion & Yes & \texttt{[VOID]} & \underline{78.32} & \textbf{39.60} & \underline{42.07} & \textbf{45.38} & \textbf{51.34} \\
\bottomrule
\end{tabular*}
\caption{Task quality under a short initial canvas. Mean arithmetically
averages the four benchmark scores.}
\end{subtable}

\vspace{1em}

\begin{subtable}[t]{\textwidth}
\centering
\begin{tabular*}{0.96\textwidth}{@{\extracolsep{\fill}}lrrrrrr@{}}
\toprule
Method & Mean Acc. \(\uparrow\) & \shortstack{GSM8K\\NFE/\textcolor{wallSpeedBlue}{Wall}} & \shortstack{MATH\\NFE/\textcolor{wallSpeedBlue}{Wall}} & \shortstack{HEval\\NFE/\textcolor{wallSpeedBlue}{Wall}} & \shortstack{MBPP\\NFE/\textcolor{wallSpeedBlue}{Wall}} & \shortstack{Mean\\NFE/\textcolor{wallSpeedBlue}{Wall}} \\
\midrule
\textsc{Daedal} & \underline{50.77} & 135.54 / \underline{\wallspd{1.00}} & \underline{269.65} / \underline{\wallspd{1.00}} & 254.66 / \wallspd{1.00} & 255.43 / \wallspd{1.00} & 228.82 / \wallspd{1.00} \\
\(\rho\)-\texttt{[EOS]} & 45.50 & \textbf{92.57} / \textbf{\wallspd{1.56}} & \textbf{104.30} / \textbf{\wallspd{3.04}} & \underline{191.52} / \underline{\wallspd{1.34}} & \underline{170.41} / \underline{\wallspd{1.69}} & \textbf{139.70} / \underline{\wallspd{1.81}} \\
VoidExpansion & \textbf{51.34} & \underline{133.78} / \wallspd{0.92} & 418.05 / \wallspd{0.82} & \textbf{73.06} / \textbf{\wallspd{5.59}} & \textbf{63.50} / \textbf{\wallspd{4.80}} & \underline{172.10} / \textbf{\wallspd{2.12}} \\
\bottomrule
\end{tabular*}
\caption{Efficiency of adaptive length selection. Each task cell reports NFE /
wall speedup. Wall speedup is normalized to \textsc{Daedal}. In the Mean
column, NFE is the arithmetic mean and wall speedup is the geometric mean over
the four benchmarks.}
\end{subtable}

\caption{Short canvas expansion results on LLaDA. VoidExpansion uses \(\tau_{\mathrm{nonvoid}}=0.09\),
\(\tau_{\mathrm{gap}}=-0.45\).}
\label{tab:gen-length-robustness}
\end{table*}

  \noindent\textbf{VoidExpansion preserves quality while improving efficiency.}
  Table~\ref{tab:gen-length-robustness} shows that
  fixed-\(L_0\) decoding reaches only 39.38 mean accuracy for the original LLaDA
  and 38.61 for the \texttt{[VOID]}-padding checkpoint.
  Thus, VoidPadding alone does not improve short-canvas decoding.
  The benefit appears when the learned \texttt{[VOID]} signal is used for
  expansion. The full VoidPadding+VoidExpansion method
  slightly improves over \textsc{Daedal} in mean quality (51.34 vs. 50.77), while
  lowering mean NFE from 228.82 to 172.10 and
  running \(2.12\times\) faster in geometric-mean wall time.
  VoidExpansion also achieves higher quality than
  \(\rho\)-\texttt{[EOS]} on all benchmarks while attaining a larger
  geometric-mean wall-clock speedup. Raw wall-clock results are shown in
  Figure~\ref{fig:appendix-expansion-efficiency-breakdown}.
  Table~\ref{tab:dream-init64-expansion} shows the same pattern on Dream:
  VoidExpansion achieves the best score on all four benchmarks, improving
  the mean score over fixed-\(L_0\) Dream (42.23) and fixed-\(L_0\)
  VoidPadding (39.20) to 60.37.

\begin{table}[t]
\centering
\scriptsize
\setlength{\tabcolsep}{3pt}
\resizebox{\columnwidth}{!}{
\begin{tabular}{lrrrrr}
\toprule
Method & GSM8K & MATH500 & HEval & MBPP & Mean \\
\midrule
Dream fixed-64 & \underline{53.22} & \underline{22.80} & \underline{43.90} & 49.00 & \underline{42.23} \\
VOID fixed-64 & 42.53 & 19.00 & 42.68 & \underline{52.60} & 39.20 \\
VoidExpansion & \textbf{79.08} & \textbf{44.00} & \textbf{60.98} & \textbf{57.40} & \textbf{60.37} \\
\bottomrule
\end{tabular}
}
\caption{Short canvas expansion results on Dream. VoidExpansion uses \(\tau_{\mathrm{nonvoid}}=0.15\),
\(\tau_{\mathrm{gap}}=0.55\).}
\label{tab:dream-init64-expansion}
\end{table}

\begin{figure}[t]
\centering
\definecolor{stressRho}{HTML}{D97706}
\definecolor{stressDaedal}{HTML}{15803D}
\definecolor{stressVoid}{HTML}{2563EB}
\newcommand{\stressscorebar}[4]{
  \pgfmathsetmacro{\h}{#3/20}
  \path[fill=#4, draw=none] (#1-0.085,0) rectangle (#1+0.085,\h);
  \node[font=\tiny, rotate=90, anchor=west, text=black] at (#1,\h+0.03) {#2};
}
\newcommand{\stressscoregroup}[5]{
  \node[font=\scriptsize, anchor=north] at (#1+0.20,-0.18) {#2};
  #3#4#5
}
\resizebox{\columnwidth}{!}{
\begin{tikzpicture}[x=1cm,y=0.64cm]
  \draw[gray!45] (0,0) -- (7.55,0);
  \draw[gray!45] (0,0) -- (0,4.55);
  \foreach \y/\lab in {0/0,1/20,2/40,3/60,4/80} {
    \draw[gray!18] (0,\y) -- (7.55,\y);
    \node[font=\tiny, anchor=east, text=black] at (-0.07,\y) {\lab};
  }
  \node[font=\scriptsize, rotate=90, anchor=center] at (-0.58,2.25) {Score};

  \stressscoregroup{0.42}{GSM8K}
    {\stressscorebar{0.42}{61.26}{61.26}{stressRho}}
    {\stressscorebar{0.62}{13.12}{13.12}{stressDaedal}}
    {\stressscorebar{0.82}{77.26}{77.26}{stressVoid}}
  \stressscoregroup{1.92}{HumanEval}
    {\stressscorebar{1.92}{35.37}{35.37}{stressRho}}
    {\stressscorebar{2.12}{34.76}{34.76}{stressDaedal}}
    {\stressscorebar{2.32}{39.63}{39.63}{stressVoid}}
  \stressscoregroup{3.42}{MATH500}
    {\stressscorebar{3.42}{23.40}{23.40}{stressRho}}
    {\stressscorebar{3.62}{12.80}{12.80}{stressDaedal}}
    {\stressscorebar{3.82}{38.60}{38.60}{stressVoid}}
  \stressscoregroup{4.92}{MBPP}
    {\stressscorebar{4.92}{43.74}{43.74}{stressRho}}
    {\stressscorebar{5.12}{33.16}{33.16}{stressDaedal}}
    {\stressscorebar{5.32}{44.66}{44.66}{stressVoid}}
  \stressscoregroup{6.42}{Mean}
    {\stressscorebar{6.42}{40.94}{40.94}{stressRho}}
    {\stressscorebar{6.62}{23.46}{23.46}{stressDaedal}}
    {\stressscorebar{6.82}{50.04}{50.04}{stressVoid}}

  \node[font=\scriptsize, anchor=center] at (3.78,4.94) {
    \tikz{\path[fill=stressRho, draw=none] (0,0) rectangle (0.24,0.10);} \(\rho\)-\texttt{[EOS]}
    \hspace{1.0em}
    \tikz{\path[fill=stressDaedal, draw=none] (0,0) rectangle (0.24,0.10);} \textsc{Daedal}
    \hspace{1.0em}
    \tikz{\path[fill=stressVoid, draw=none] (0,0) rectangle (0.24,0.10);} VoidExpansion
  };
\end{tikzpicture}
}
\caption{Large-block stress test with \(L_0=256\) and \(B=256\). Lower scores indicate \texttt{[EOS]} overflow.}
\label{fig:gsm8k-large-block-eos-stress}
\end{figure}
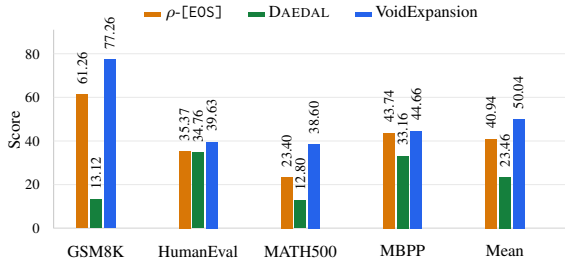

  To test large-block robustness, we set \(L_0=256\) and \(B=256\). We apply
  \(\rho\)-\texttt{[EOS]} and \textsc{Daedal} to the original LLaDA model, and
  compare them with VoidExpansion on the VoidPadding checkpoint.
  Figure~\ref{fig:gsm8k-large-block-eos-stress} shows that \textsc{Daedal}
  collapses to a 23.46 mean score, including 13.12 on GSM8K and 12.80 on
  MATH500. \(\rho\)-\texttt{[EOS]} is more stable but still underperforms
  VoidExpansion on every benchmark, with a 9.10-point lower mean score
  (40.94 vs. 50.04). This confirms that inference-only methods relying on the
  original \texttt{[EOS]} signal remain vulnerable to large-block
  \texttt{[EOS]} overflow.

\let\stressscorebar\relax
\let\stressscoregroup\relax

We further evaluate the effectiveness of VoidPadding and VoidExpansion on
LLaDA-8B-Base and Dream-7B-Base in
Appendices~\ref{app:base-model-blocksize-sweeps}
and~\ref{app:base-void-expansion-threshold-sweeps}.

\section{Ablation Studies}
\label{sec:ablations}

\noindent\textbf{Setup.} 
 Our ablation experiments use the same benchmarks and metrics as Section~\ref{sec:experiments}
  and compare the original LLaDA-8B-Instruct model with the RainbowPadding and VoidPadding checkpoints.

  \noindent\textbf{VoidPadding makes \texttt{[EOS]}Termination effective.}
  Table~\ref{tab:training-ablation-block-sweep} evaluates \texttt{[EOS]}Termination on a fixed \(512\)-token canvas across different training settings.
  Relative to the VanillaStopping baseline defined in
  Section~\ref{subsec:void-padding}, \texttt{[EOS]}Termination reduces NFE for LLaDA and RainbowPadding but slightly degrades
  accuracy.  In
  contrast, with VoidPadding, \texttt{[EOS]}Termination improves
  accuracy while reducing mean NFE by 63.8\% across block sizes.  This
  improvement indicates that VoidPadding aligns training and
  inference so that \texttt{[EOS]} becomes a reliable semantic terminator for
  early stopping. Per-benchmark NFE results are shown in
  Figure~\ref{fig:appendix-fixed-length-nfe-by-benchmark}.

\begin{table}[t]
\centering
\scriptsize
\setlength{\tabcolsep}{3pt}
\resizebox{\columnwidth}{!}{
\begin{tabular}{lcccc}
\toprule
\multirow{2}{*}{Setting} &
\multicolumn{3}{c}{Mean score} &
\multirow{2}{*}{Mean NFE reduction} \\
\cmidrule(lr){2-4}
& VanillaStopping & \texttt{[EOS]}Termination & \(\Delta\) & \\
\midrule
LLaDA & 43.28 & 43.15 & -0.13 & 15.3\% \\
RainbowPadding & 46.75 & 46.74 & -0.01 & 44.7\% \\
VoidPadding & 40.70 & 50.34 & +9.64 & 63.8\% \\
\bottomrule
\end{tabular}
}
\caption{\texttt{[EOS]}Termination ablation on LLaDA. Mean scores    and mean NFE are averaged over the four benchmarks and \(B\in\{64,128,512\}\). NFE reduction is relative to VanillaStopping.}
\label{tab:training-ablation-block-sweep}
\end{table}

\definecolor{voidsupportred}{RGB}{255,80,80}
\definecolor{voidsupportgreen}{RGB}{0,190,120}

\newcommand{\VoidTauHeatCell}[5]{
  \pgfmathsetmacro{\voidnorm}{(#3-#4)/(#5-#4)}
  \ifdim \voidnorm pt < 0.5pt
    \pgfmathtruncatemacro{\voidshade}{100*(1-2*\voidnorm)}
    \fill[voidsupportred!\voidshade!white, draw=black!65, line width=0.22pt] (#1,#2) rectangle ++(1,1);
  \else
    \pgfmathtruncatemacro{\voidshade}{100*(2*\voidnorm-1)}
    \fill[voidsupportgreen!\voidshade!white, draw=black!65, line width=0.22pt] (#1,#2) rectangle ++(1,1);
  \fi
}
\newcommand{\VoidTauPanelFrame}[1]{
  \node[font=\scriptsize\bfseries] at (2,4.74) {#1};
  \node[font=\tiny] at (2,4.38) {\(\tau_{\mathrm{nonvoid}}\)};
  \foreach \x/\lab in {0.5/0.06,1.5/0.09,2.5/0.12,3.5/0.15} {
    \node[font=\tiny] at (\x,4.14) {\lab};
  }
  \foreach \y/\lab in {3.5/-0.55,2.5/-0.50,1.5/-0.45,0.5/-0.40} {
    \node[anchor=east,font=\tiny] at (-0.14,\y) {\lab};
  }
  \node[rotate=90,font=\tiny] at (-0.62,2) {\(\tau_{\mathrm{gap}}\)};
  \draw[line width=0.35pt, black!55] (0,0) rectangle (4,4);
  \draw[blue!80!black, line width=0.9pt] (1,2) rectangle (2,3);
}
\newcommand{\VoidTauColorScale}[2]{
  \foreach \i in {0,...,19} {
    \pgfmathsetmacro{\voidbarnorm}{\i/19}
    \ifdim \voidbarnorm pt < 0.5pt
      \pgfmathtruncatemacro{\voidbarshade}{100*(1-2*\voidbarnorm)}
      \fill[voidsupportred!\voidbarshade!white] (4.12,{0.20*\i}) rectangle (4.48,{0.20*(\i+1)});
    \else
      \pgfmathtruncatemacro{\voidbarshade}{100*(2*\voidbarnorm-1)}
      \fill[voidsupportgreen!\voidbarshade!white] (4.12,{0.20*\i}) rectangle (4.48,{0.20*(\i+1)});
    \fi
  }
  \draw[line width=0.45pt, black!65] (4.12,0) rectangle (4.48,4);
  \node[anchor=north,font=\scriptsize\bfseries] at (4.30,-0.08) {#1};
  \node[anchor=south,font=\scriptsize\bfseries] at (4.30,4.08) {#2};
}
\newcommand{\VoidTauBaselineMark}[5]{
  \pgfmathsetmacro{\voidmarky}{4*(#1-#3)/(#4-#3)}
  \draw[#5, line width=1.45pt] (4.03,\voidmarky) -- (4.58,\voidmarky);
  \node[anchor=west,font=\small\bfseries,text=#5] at (4.65,\voidmarky) {#2};
}
\newcommand{\VoidTauInstructHeatCell}[5]{
  \pgfmathsetmacro{\voidnorm}{(#3-#4)/(#5-#4)}
  \ifdim \voidnorm pt < 0.5pt
    \pgfmathtruncatemacro{\voidshade}{100*(1-2*\voidnorm)}
    \fill[voidsupportred!\voidshade!white, draw=black!65, line width=0.22pt] (#1,#2) rectangle ++(1,1);
  \else
    \pgfmathtruncatemacro{\voidshade}{100*(2*\voidnorm-1)}
    \fill[voidsupportgreen!\voidshade!white, draw=black!65, line width=0.22pt] (#1,#2) rectangle ++(1,1);
  \fi
}
\newcommand{\VoidTauInstructColorScale}[2]{
  \foreach \i in {0,...,19} {
    \pgfmathsetmacro{\voidbarnorm}{\i/19}
    \ifdim \voidbarnorm pt < 0.5pt
      \pgfmathtruncatemacro{\voidbarshade}{100*(1-2*\voidbarnorm)}
      \fill[voidsupportred!\voidbarshade!white] (4.12,{0.20*\i}) rectangle (4.48,{0.20*(\i+1)});
    \else
      \pgfmathtruncatemacro{\voidbarshade}{100*(2*\voidbarnorm-1)}
      \fill[voidsupportgreen!\voidbarshade!white] (4.12,{0.20*\i}) rectangle (4.48,{0.20*(\i+1)});
    \fi
  }
  \draw[line width=0.45pt, black!65] (4.12,0) rectangle (4.48,4);
  \node[anchor=north,font=\scriptsize\bfseries] at (4.30,-0.08) {#1};
  \node[anchor=south,font=\scriptsize\bfseries] at (4.30,4.08) {#2};
}
\newcommand{\VoidTauInstructBaselineMark}[5]{
  \pgfmathsetmacro{\voidmarky}{4*(#1-#3)/(#4-#3)}
  \draw[#5, line width=1.45pt] (4.03,\voidmarky) -- (4.58,\voidmarky);
  \node[anchor=west,font=\small\bfseries,text=#5] at (4.65,\voidmarky) {#2};
}
\newcommand{\VoidTauInstructPanelFrame}[1]{
  \node[font=\scriptsize\bfseries] at (2,4.74) {#1};
  \node[font=\tiny] at (2,4.38) {\(\tau_{\mathrm{nonvoid}}\)};
  \foreach \x/\lab in {0.5/0.03,1.5/0.06,2.5/0.09,3.5/0.12} {
    \node[font=\tiny] at (\x,4.14) {\lab};
  }
  \foreach \y/\lab in {3.5/-0.45,2.5/-0.40,1.5/-0.35,0.5/-0.30} {
    \node[anchor=east,font=\tiny] at (-0.14,\y) {\lab};
  }
  \node[rotate=90,font=\tiny] at (-0.62,2) {\(\tau_{\mathrm{gap}}\)};
  \draw[line width=0.35pt, black!55] (0,0) rectangle (4,4);
  \draw[blue!80!black, line width=0.9pt] (2,3) rectangle (3,4);
}

\noindent\textbf{VoidExpansion is robust to hyperparameter choices.}
Starting from the VoidExpansion setting in
Table~\ref{tab:gen-length-robustness},
Table~\ref{tab:void-expansion-hparam-summary} varies one factor at a time and
keeps the remaining settings fixed.  The varied factors include threshold
pairs, initial lengths $L_0$, tail window sizes $w$, and block sizes $B$.  Mean quality and
mean NFE over the four tasks remain stable across these changes, showing
that the learned \texttt{[VOID]} length signal supports short-canvas expansion
while preserving both quality and efficiency. 
 See Appendix~\ref{app:void-expansion-threshold-sweeps} for detailed results behind Table~\ref{tab:void-expansion-hparam-summary}.

\begin{table}[t]
\centering
\scriptsize
\setlength{\tabcolsep}{3pt}
\resizebox{\columnwidth}{!}{
\begin{tabular}{llcc}
\toprule
Factor varied & Values tested & Task-mean quality range & Task-mean NFE range \\
\midrule
Thresholds & 16 threshold pairs (\(4{\times}4\)) & 51.16--51.78 & 169.39--175.50 \\
Initial length \(L_0\) & 64, 96, 128, 160 & 50.83--51.78 & 172.10--172.80 \\
Tail window size \(w\) & 8, 16, 24, 32 & 51.40--51.81 & 168.65--173.15 \\
Block size \(B\) & 16, 32, 64 & 50.80--51.78 & 170.74--172.35 \\
\bottomrule
\end{tabular}
}
\caption{Hyperparameter robustness of VoidExpansion. Each row varies one factor from the setting in Table~\ref{tab:gen-length-robustness} while keeping the others fixed. Quality and NFE are averaged over the four benchmarks. Each range shows the four-benchmark mean quality or NFE obtained across the values of that factor listed in the row.}
\label{tab:void-expansion-hparam-summary}
\end{table}

\section{Related Work}

\noindent\textbf{Diffusion Language Models.} 
Diffusion language models (DLMs) can be broadly categorized into continuous DLMs~\citep{guo2026continuous,hu2026elf,jo2026continuous,shen2026codar}, which operate in embedding spaces, and discrete DLMs~\citep{hoogeboom2021argmax,sahoo2025diffusion,do2025discrete}, which operate over discrete token spaces. 
 Masked diffusion language models~\citep{hersche2025soft,he2023diffusionbert,austin2021structured} (MDLMs) are a representative discrete class that iteratively predicts masked tokens.
 Recent large-scale MDLMs, including LLaDA series~\cite{nie2026large, zhu2025llada, bie2025llada2, bie2026llada2}, Dream series~\cite{ye2025dream, wu2026dreamon}, SDAR~\cite{cheng2025sdar}, TraDo~\cite{wang2025revolutionizing}, and WeDLM~\cite{liu2025wedlm}, show strong complex reasoning and long-text generation abilities.
 Recent work also improves generation quality through better samplers~\citep{yang2026improving}.
  
\noindent\textbf{Efficient MDLMs.}
Although MDLMs support parallel token prediction, inference remains costly because generation usually requires many denoising iterations with full-sequence bidirectional attention. Work on efficient MDLM inference includes sampling methods that reduce denoising steps or finalize more tokens per step~\citep{li2025diffusion,israel2026accelerating,wei2025accelerating}, cache-based methods that reuse computation via approximate caching or KV caching~\cite{wu2025fast, hu2025flashdlm, song2026sparse, wudynamic, luo2026dsb,wei2025orchestrating}, and training-based distillation for fewer-step generation~\citep{monsefi2025fs, amin2026consistent, chen2025dlm}.

\noindent\textbf{Variable-Length Generation.} 
MDLMs often require a preallocated output canvas, wasting computation on over-allocated slots and limiting usability~\citep{han2023ssd}. Length allocation methods address this issue~\cite{cheng2026improving}. Inference-only methods, including SmartCrop~\cite{rossi2026diffusion}, \(\rho\)-\texttt{[EOS]}~\cite{yang2026rho}, and DAEDAL~\cite{li2025beyond}, typically use \texttt{[EOS]} as a length signal, while training-based methods include dLLM-Var~\citep{yang2025diffusion}, which learns reliable \texttt{[EOS]} prediction for block-wise termination, and FlexMDM~\citep{kim2025any}, which decouples length prediction from token prediction.

\section{Conclusion}

We show that a root cause of \texttt{[EOS]} overflow in MDLMs is a
  training--inference role mismatch: \texttt{[EOS]} is trained as both padding
  and semantic terminator, but used only for termination at inference.
  VoidPadding instead uses \texttt{[VOID]} as padding tokens and reserves
  \texttt{[EOS]} for semantic endings, enabling VoidExpansion for canvas
  allocation and \texttt{[EOS]}Termination for early stopping.
  Together, our training--inference co-design improves large-block robustness,
  decoding efficiency, and variable-length generation.

 \section*{Limitations}
  Many MDLM training and inference conventions are inherited from ARLMs.
  Our study focuses on one concrete failure mode of this transfer:
  \texttt{[EOS]} padding, which is largely benign under causal masking, can
  become fragile under bidirectional denoising because padding positions can
  serve as both context and denoising targets.
  Thus, \texttt{[EOS]} overflow should be viewed as one symptom of a broader
  class of ARLM-inherited assumptions that may fail in MDLMs.
  Other choices, including data formatting, special-token semantics, stopping
  rules, and batching strategies, may require similarly MDLM-specific design.

Our analysis also suggests that \texttt{[EOS]} overflow is not primarily a model
capacity failure. Instead, it emerges from the interaction between a learned
length prior, bidirectional context, and standard large-block sampling rules.
  VoidPadding mitigates the padding--termination mismatch, but the design of
  effective MDLM samplers remains open.
  Future samplers may need to jointly consider token confidence, commitment
  timing, canvas allocation, and stopping decisions under bidirectional denoising.

\bibliography{custom}

@article{yang2025diffusion,
  title={Diffusion llm with native variable generation lengths: Let [eos] lead the way},
  author={Yang, Yicun and Wang, Cong and Wang, Shaobo and Wen, Zichen and Qi, Biqing and Xu, Hanlin and Zhang, Linfeng},
  journal={arXiv preprint arXiv:2510.24605},
  year={2025}
}

@article{chung2024scaling,
  title={Scaling instruction-finetuned language models},
  author={Chung, Hyung Won and Hou, Le and Longpre, Shayne and Zoph, Barret and Tay, Yi and Fedus, William and Li, Yunxuan and Wang, Xuezhi and Dehghani, Mostafa and Brahma, Siddhartha and others},
  journal={Journal of Machine Learning Research},
  volume={25},
  number={70},
  pages={1--53},
  year={2024}
}

@article{kim2025rainbow,
  title={Rainbow Padding: Mitigating Early Termination in Instruction-Tuned Diffusion LLMs},
  author={Kim, Bumjun and Jeon, Dongjae and Kim, Dueun and Jeung, Wonje and No, Albert},
  journal={arXiv preprint arXiv:2510.03680},
  year={2025}
}

@article{vaswani2017attention,
  title={Attention is all you need},
  author={Vaswani, Ashish and Shazeer, Noam and Parmar, Niki and Uszkoreit, Jakob and Jones, Llion and Gomez, Aidan N and Kaiser, {\L}ukasz and Polosukhin, Illia},
  journal={Advances in neural information processing systems},
  volume={30},
  year={2017}
}

@article{nie2026large,
  title={Large language diffusion models},
  author={Nie, Shen and Zhu, Fengqi and You, Zebin and Zhang, Xiaolu and Ou, Jingyang and Hu, Jun and Zhou, Jun and Lin, Yankai and Wen, Ji-Rong and Li, Chongxuan},
  journal={Advances in Neural Information Processing Systems},
  volume={38},
  pages={50608--50646},
  year={2026}
}

@article{li2025beyond,
  title={Beyond fixed: Training-free variable-length denoising for diffusion large language models},
  author={Li, Jinsong and Dong, Xiaoyi and Zang, Yuhang and Cao, Yuhang and Wang, Jiaqi and Lin, Dahua},
  journal={arXiv preprint arXiv:2508.00819},
  year={2025}
}

@article{yang2026rho,
  title={{{\ensuremath{\rho}}-EOS}: Training-free Bidirectional Variable-Length Control for Masked Diffusion LLMs},
  author={Yang, Jingyi and Jiang, Yuxian and Shao, Jing},
  journal={arXiv preprint arXiv:2601.22527},
  year={2026}
}

@article{rossi2026diffusion,
  title={Diffusion Language Models Are Natively Length-Aware},
  author={Rossi, Vittorio and Cir{\`o}, Giacomo and Beltrame, Davide and Gandolfi, Luca and R{\"o}ttger, Paul and Hovy, Dirk},
  journal={arXiv preprint arXiv:2603.06123},
  year={2026}
}

@article{austin2021structured,
  title={Structured denoising diffusion models in discrete state-spaces},
  author={Austin, Jacob and Johnson, Daniel D and Ho, Jonathan and Tarlow, Daniel and Van Den Berg, Rianne},
  journal={Advances in neural information processing systems},
  volume={34},
  pages={17981--17993},
  year={2021}
}

@article{sahoo2024simple,
  title={Simple and effective masked diffusion language models},
  author={Sahoo, Subham S and Arriola, Marianne and Schiff, Yair and Gokaslan, Aaron and Marroquin, Edgar and Chiu, Justin T and Rush, Alexander and Kuleshov, Volodymyr},
  journal={Advances in Neural Information Processing Systems},
  volume={37},
  pages={130136--130184},
  year={2024}
}

@article{ye2025dream,
  title={Dream 7b: Diffusion large language models},
  author={Ye, Jiacheng and Xie, Zhihui and Zheng, Lin and Gao, Jiahui and Wu, Zirui and Jiang, Xin and Li, Zhenguo and Kong, Lingpeng},
  journal={arXiv preprint arXiv:2508.15487},
  year={2025}
}

@article{bie2026llada2,
  title={Llada2. 1: Speeding up text diffusion via token editing},
  author={Bie, Tiwei and Cao, Maosong and Cao, Xiang and Chen, Bingsen and Chen, Fuyuan and Chen, Kun and Du, Lun and Feng, Daozhuo and Feng, Haibo and Gong, Mingliang and others},
  journal={arXiv preprint arXiv:2602.08676},
  year={2026}
}

@article{radford2018improving,
  title={Improving language understanding by generative pre-training},
  author={Radford, Alec and Narasimhan, Karthik and Salimans, Tim and Sutskever, Ilya and others},
  year={2018},
  publisher={San Francisco, CA, USA}
}

@article{wu2025fast,
  title={Fast-dllm: Training-free acceleration of diffusion llm by enabling kv cache and parallel decoding},
  author={Wu, Chengyue and Zhang, Hao and Xue, Shuchen and Liu, Zhijian and Diao, Shizhe and Zhu, Ligeng and Luo, Ping and Han, Song and Xie, Enze},
  journal={arXiv preprint arXiv:2505.22618},
  year={2025}
}

@article{yang2026improving,
  title={Improving Sampling for Masked Diffusion Models via Information Gain},
  author={Yang, Kaisen and Teoh, Jayden and Yang, Kaicheng and Zhang, Yitong and Lamb, Alex},
  journal={arXiv preprint arXiv:2602.18176},
  year={2026}
}

@article{cheng2026improving,
  title={Improving Variable-Length Generation in Diffusion Language Models via Length Regularization},
  author={Cheng, Zicong and Jia, Ruixuan and Li, Jia and Yang, Guo-Wei and Guo, Meng-Hao and Hu, Shi-Min},
  journal={arXiv preprint arXiv:2602.07546},
  year={2026}
}

@article{ben2026accelerated,
  title={Accelerated sampling from masked diffusion models via entropy bounded unmasking},
  author={Ben-Hamu, Heli and Gat, Itai and Severo, Daniel and Nolte, Niklas S and Karrer, Brian},
  journal={Advances in Neural Information Processing Systems},
  volume={38},
  pages={55981--56007},
  year={2026}
}

@article{hu2025flashdlm,
  title={FlashDLM: Accelerating Diffusion Language Model Inference via Efficient KV Caching and Guided Diffusion},
  author={Hu, Zhanqiu and Meng, Jian and Akhauri, Yash and Abdelfattah, Mohamed S and Seo, Jae-sun and Zhang, Zhiru and Gupta, Udit},
  journal={arXiv preprint arXiv:2505.21467},
  year={2025}
}

@article{wu2026dreamon,
  title={DreamOn: Diffusion Language Models For Code Infilling Beyond Fixed-size Canvas},
  author={Wu, Zirui and Zheng, Lin and Xie, Zhihui and Ye, Jiacheng and Gao, Jiahui and Gong, Shansan and Feng, Yansong and Li, Zhenguo and Bi, Wei and Zhou, Guorui and others},
  journal={arXiv preprint arXiv:2602.01326},
  year={2026}
}

@article{austin2021program,
  title={Program synthesis with large language models},
  author={Austin, Jacob and Odena, Augustus and Nye, Maxwell and Bosma, Maarten and Michalewski, Henryk and Dohan, David and Jiang, Ellen and Cai, Carrie and Terry, Michael and Le, Quoc and others},
  journal={arXiv preprint arXiv:2108.07732},
  year={2021}
}

@article{chen2021evaluating,
  title={Evaluating large language models trained on code},
  author={Chen, Mark and Tworek, Jerry and Jun, Heewoo and Yuan, Qiming and Pinto, Henrique Ponde De Oliveira and Kaplan, Jared and Edwards, Harri and Burda, Yuri and Joseph, Nicholas and Brockman, Greg and others},
  journal={arXiv preprint arXiv:2107.03374},
  year={2021}
}

@article{hendrycks2021measuring,
  title={Measuring mathematical problem solving with the math dataset},
  author={Hendrycks, Dan and Burns, Collin and Kadavath, Saurav and Arora, Akul and Basart, Steven and Tang, Eric and Song, Dawn and Steinhardt, Jacob},
  journal={arXiv preprint arXiv:2103.03874},
  year={2021}
}

@article{cobbe2021training,
  title={Training verifiers to solve math word problems},
  author={Cobbe, Karl and Kosaraju, Vineet and Bavarian, Mohammad and Chen, Mark and Jun, Heewoo and Kaiser, Lukasz and Plappert, Matthias and Tworek, Jerry and Hilton, Jacob and Nakano, Reiichiro and others},
  journal={arXiv preprint arXiv:2110.14168},
  year={2021}
}

@article{lambert2024tulu,
  title={Tulu 3: Pushing frontiers in open language model post-training},
  author={Lambert, Nathan and Morrison, Jacob and Pyatkin, Valentina and Huang, Shengyi and Ivison, Hamish and Brahman, Faeze and Miranda, Lester James V and Liu, Alisa and Dziri, Nouha and Lyu, Shane and others},
  journal={arXiv preprint arXiv:2411.15124},
  year={2024}
}

@article{allal2024smollm,
  title={SmolLM-blazingly fast and remarkably powerful},
  author={Allal, Loubna Ben and Lozhkov, Anton and Bakouch, Elie and von Werra, Leandro and Wolf, Thomas},
  journal={Hugging Face Blog},
  volume={16},
  year={2024}
}

@article{lou2023discrete,
  title={Discrete diffusion modeling by estimating the ratios of the data distribution},
  author={Lou, Aaron and Meng, Chenlin and Ermon, Stefano},
  journal={arXiv preprint arXiv:2310.16834},
  year={2023}
}

@article{kim2025train,
  title={Train for the worst, plan for the best: Understanding token ordering in masked diffusions},
  author={Kim, Jaeyeon and Shah, Kulin and Kontonis, Vasilis and Kakade, Sham and Chen, Sitan},
  journal={arXiv preprint arXiv:2502.06768},
  year={2025}
}

@inproceedings{arriola2025block,
  title={Block diffusion: Interpolating between autoregressive and diffusion language models},
  author={Arriola, Marianne and Gokaslan, Aaron and Chiu, Justin and Yang, Zhihan and Qi, Zhixuan and Han, Jiaqi and Sahoo, Subham and Kuleshov, Volodymyr},
  booktitle={International Conference on Learning Representations},
  volume={2025},
  pages={50726--50753},
  year={2025}
}

@article{cheng2025sdar,
  title={Sdar: A synergistic diffusion-autoregression paradigm for scalable sequence generation},
  author={Cheng, Shuang and Bian, Yihan and Liu, Dawei and Zhang, Linfeng and Yao, Qian and Tian, Zhongbo and Wang, Wenhai and Guo, Qipeng and Chen, Kai and Qi, Biqing and others},
  journal={arXiv preprint arXiv:2510.06303},
  year={2025}
}

@article{wang2025revolutionizing,
  title={Revolutionizing reinforcement learning framework for diffusion large language models},
  author={Wang, Yinjie and Yang, Ling and Li, Bowen and Tian, Ye and Shen, Ke and Wang, Mengdi},
  journal={arXiv preprint arXiv:2509.06949},
  year={2025}
}

@article{cai2026confidence,
  title={Confidence-Based Decoding is Provably Efficient for Diffusion Language Models},
  author={Cai, Changxiao and Li, Gen},
  journal={arXiv preprint arXiv:2603.22248},
  year={2026}
}

@article{hu2022lora,
  title={Lora: Low-rank adaptation of large language models.},
  author={Hu, Edward J and Shen, Yelong and Wallis, Phillip and Allen-Zhu, Zeyuan and Li, Yuanzhi and Wang, Shean and Wang, Liang and Chen, Weizhu and others},
  journal={Iclr},
  volume={1},
  number={2},
  pages={3},
  year={2022}
}

@article{hersche2025soft,
  title={Soft-masked diffusion language models},
  author={Hersche, Michael and Moor-Smith, Samuel and Hofmann, Thomas and Rahimi, Abbas},
  journal={arXiv preprint arXiv:2510.17206},
  year={2025}
}

@article{zhu2025llada,
  title={Llada 1.5: Variance-reduced preference optimization for large language diffusion models},
  author={Zhu, Fengqi and Wang, Rongzhen and Nie, Shen and Zhang, Xiaolu and Wu, Chunwei and Hu, Jun and Zhou, Jun and Chen, Jianfei and Lin, Yankai and Wen, Ji-Rong and others},
  journal={arXiv preprint arXiv:2505.19223},
  year={2025}
}

@article{bie2025llada2,
  title={Llada2. 0: Scaling up diffusion language models to 100b},
  author={Bie, Tiwei and Cao, Maosong and Chen, Kun and Du, Lun and Gong, Mingliang and Gong, Zhuochen and Gu, Yanmei and Hu, Jiaqi and Huang, Zenan and Lan, Zhenzhong and others},
  journal={arXiv preprint arXiv:2512.15745},
  year={2025}
}

@inproceedings{song2026sparse,
  title={Sparse-dllm: Accelerating diffusion llms with dynamic cache eviction},
  author={Song, Yuerong and Liu, Xiaoran and Li, Ruixiao and Liu, Zhigeng and Huang, Zengfeng and Guo, Qipeng and He, Ziwei and Qiu, Xipeng},
  booktitle={Proceedings of the AAAI Conference on Artificial Intelligence},
  volume={40},
  number={39},
  pages={33038--33046},
  year={2026}
}

@inproceedings{wudynamic,
  title={Dynamic-dLLM: Dynamic Cache-Budget and Adaptive Parallel Decoding for Training-Free Acceleration of Diffusion LLM},
  author={Wu, Tianyi and Sun, Xiaoxi and Jiao, Yanhua and Li, Yulin and Chen, Yixin and Cao, Yun-Hao and Hu, Yi-Qi and Tian, Zhuotao},
  booktitle={The Fourteenth International Conference on Learning Representations},
  year={2026}
}

@article{luo2026dsb,
  title={DSB: Dynamic Sliding Block Scheduling for Diffusion LLMs},
  author={Luo, Lizhuo and Li, Shenggui and Wen, Yonggang and Zhang, Tianwei},
  journal={arXiv preprint arXiv:2602.05992},
  year={2026}
}

@article{hu2026elf,
  title={ELF: Embedded Language Flows},
  author={Hu, Keya and Qiu, Linlu and Lu, Yiyang and Zhao, Hanhong and Li, Tianhong and Kim, Yoon and Andreas, Jacob and He, Kaiming},
  journal={arXiv preprint arXiv:2605.10938},
  year={2026}
}

@article{liu2025wedlm,
  title={Wedlm: Reconciling diffusion language models with standard causal attention for fast inference},
  author={Liu, Aiwei and He, Minghua and Zeng, Shaoxun and Zhang, Sijun and Zhang, Linhao and Wu, Chuhan and Jia, Wei and Liu, Yuan and Zhou, Xiao and Zhou, Jie},
  journal={arXiv preprint arXiv:2512.22737},
  year={2025}
}

@article{guo2026continuous,
  title={Continuous Latent Diffusion Language Model},
  author={Guo, Hongcan and Zhao, Qinyu and Zhao, Yian and Nie, Shen and Zhu, Rui and Guo, Qiushan and Wang, Feng and Yang, Tao and Zhao, Hengshuang and Wei, Guoqiang and others},
  journal={arXiv preprint arXiv:2605.06548},
  year={2026}
}

@article{jo2026continuous,
  title={Continuous diffusion model for language modeling},
  author={Jo, Jaehyeong and Hwang, Sung Ju},
  journal={Advances in Neural Information Processing Systems},
  volume={38},
  pages={97394--97430},
  year={2026}
}

@article{shen2026codar,
  title={Codar: Continuous diffusion language models are more powerful than you think},
  author={Shen, Junzhe and Zhao, Jieru and He, Ziwei and Lin, Zhouhan},
  journal={arXiv preprint arXiv:2603.02547},
  year={2026}
}

@inproceedings{do2025discrete,
  title={Discrete diffusion language model for efficient text summarization},
  author={Do, Duc Anh and Tuan, Luu Anh and Buntine, Wray and others},
  booktitle={Findings of the Association for Computational Linguistics: NAACL 2025},
  pages={6278--6290},
  year={2025}
}

@article{sahoo2025diffusion,
  title={The diffusion duality},
  author={Sahoo, Subham Sekhar and Deschenaux, Justin and Gokaslan, Aaron and Wang, Guanghan and Chiu, Justin and Kuleshov, Volodymyr},
  journal={arXiv preprint arXiv:2506.10892},
  year={2025}
}

@article{hoogeboom2021argmax,
  title={Argmax flows and multinomial diffusion: Learning categorical distributions},
  author={Hoogeboom, Emiel and Nielsen, Didrik and Jaini, Priyank and Forr{\'e}, Patrick and Welling, Max},
  journal={Advances in neural information processing systems},
  volume={34},
  pages={12454--12465},
  year={2021}
}

@inproceedings{he2023diffusionbert,
  title={Diffusionbert: Improving generative masked language models with diffusion models},
  author={He, Zhengfu and Sun, Tianxiang and Tang, Qiong and Wang, Kuanning and Huang, Xuan-Jing and Qiu, Xipeng},
  booktitle={Proceedings of the 61st annual meeting of the association for computational linguistics (volume 1: Long papers)},
  pages={4521--4534},
  year={2023}
}

@article{li2025diffusion,
  title={Diffusion language models know the answer before decoding},
  author={Li, Pengxiang and Zhou, Yefan and Muhtar, Dilxat and Yin, Lu and Yan, Shilin and Shen, Li and Vosoughi, Soroush and Liu, Shiwei},
  journal={arXiv preprint arXiv:2508.19982},
  year={2025}
}

@article{israel2026accelerating,
  title={Accelerating diffusion llms via adaptive parallel decoding},
  author={Israel, Daniel and Van den Broeck, Guy and Grover, Aditya},
  journal={Advances in neural information processing systems},
  volume={38},
  pages={52870--52888},
  year={2026}
}

@article{wei2025accelerating,
  title={Accelerating diffusion large language models with slowfast sampling: The three golden principles},
  author={Wei, Qingyan and Zhang, Yaojie and Liu, Zhiyuan and Zeng, Puyu and Wang, Yuxuan and Qi, Biqing and Liu, Dongrui and Zhang, Linfeng},
  journal={arXiv preprint arXiv:2506.10848},
  year={2025}
}

@article{monsefi2025fs,
  title={Fs-dfm: Fast and accurate long text generation with few-step diffusion language models},
  author={Monsefi, Amin Karimi and Bhendawade, Nikhil and Ciosici, Manuel Rafael and Culver, Dominic and Zhang, Yizhe and Belousova, Irina},
  journal={arXiv preprint arXiv:2509.20624},
  year={2025}
}

@article{amin2026consistent,
  title={Consistent Diffusion Language Models},
  author={Amin, Hasan and Gao, Yuan and Souri, Yaser and Som, Subhojit and Yin, Ming and Khanna, Rajiv and Song, Xia},
  journal={arXiv preprint arXiv:2605.00161},
  year={2026}
}

@article{chen2025dlm,
  title={Dlm-one: Diffusion language models for one-step sequence generation},
  author={Chen, Tianqi and Zhang, Shujian and Zhou, Mingyuan},
  journal={arXiv preprint arXiv:2506.00290},
  year={2025}
}

@article{kim2025any,
  title={Any-order flexible length masked diffusion},
  author={Kim, Jaeyeon and Cheuk-Kit, Lee and Domingo-Enrich, Carles and Du, Yilun and Kakade, Sham and Ngotiaoco, Timothy and Chen, Sitan and Albergo, Michael},
  journal={arXiv preprint arXiv:2509.01025},
  year={2025}
}

@inproceedings{han2023ssd,
  title={Ssd-lm: Semi-autoregressive simplex-based diffusion language model for text generation and modular control},
  author={Han, Xiaochuang and Kumar, Sachin and Tsvetkov, Yulia},
  booktitle={Proceedings of the 61st Annual Meeting of the Association for Computational Linguistics (Volume 1: Long Papers)},
  pages={11575--11596},
  year={2023}
}

@article{wei2025orchestrating,
  title={Orchestrating Dual-Boundaries: An Arithmetic Intensity Inspired Acceleration Framework for Diffusion Language Models},
  author={Wei, Linye and Chen, Wenjue and Tang, Pingzhi and Guo, Xiaotian and Ye, Le and Wang, Runsheng and Li, Meng},
  journal={arXiv preprint arXiv:2511.21759},
  year={2025}
}

@inproceedings{shi2024simplified,
  title={Simplified and Generalized Masked Diffusion for Discrete Data},
  author={Shi, Jiaxin and Han, Kehang and Wang, Zhe and Doucet, Arnaud and Titsias, Michalis K.},
  booktitle={Advances in Neural Information Processing Systems},
  year={2024}
}

\appendix

\section{Training Configurations}
\label{app:training-configuration}

\paragraph{Dataset.}
We use the same 0.5M-example instruction-tuning dataset as the official
RainbowPadding~\citep{kim2025rainbow} implementation.  The data pool combines Tulu3~\citep{lambert2024tulu}
and SmolLM2~\citep{allal2024smollm}.  After removing examples longer than 4096
tokens and multi-turn dialogues, we randomly sample 0.5M examples and use this
identical data for all models in our experiments.

\paragraph{Shared training setup.}
Following the official RainbowPadding training setup, we train LLaDA-8B-Instruct~\citep{nie2026large} and Dream-7B-Instruct~\citep{ye2025dream} for the main
experiments, and use LLaDA-8B-Base and Dream-7B-Base for additional diagnostic
runs.  All models are finetuned with LoRA~\citep{hu2022lora} of rank 32 applied
to all linear layers, without bias terms.  We use AdamW with learning rate
\(5\times10^{-5}\) and global batch size 48.  LLaDA models are trained on two
A800-80GB GPUs with mini-batch size 6 per GPU and 4 gradient-accumulation
steps; Dream models are trained on three A800-80GB GPUs with mini-batch size 4
per GPU and 4 gradient-accumulation steps. For both models,  one epoch takes approximately
32--34 A800 GPU hours.

\paragraph{Training budgets.}
RainbowPadding follows its official budget: one epoch for LLaDA-8B-Instruct and
Dream-7B-Instruct, corresponding to 10,417 optimizer updates, and three epochs
for LLaDA-8B-Base and Dream-7B-Base.  All VoidPadding checkpoints use 67\% of
the corresponding RainbowPadding training budget: 7000 optimizer updates for
LLaDA-8B-Instruct and Dream-7B-Instruct, and two epochs for LLaDA-8B-Base and
Dream-7B-Base.  The \texttt{[EOS]}-padding diagnostic models are trained for two
epochs on LLaDA-8B-Base and Dream-7B-Base.

\paragraph{67\% training-budget comparison.}
Figures~\ref{tab:appendix-instruct-rainbow67-void} and
\ref{fig:dream-instruct-rainbow67-void} compare Early RainbowPadding and
VoidPadding under the same training budget.  Specifically, Early RainbowPadding
uses 7000 optimizer updates, matching the VoidPadding training budget and
corresponding to 67\% of the official RainbowPadding budget for both
LLaDA-8B-Instruct and Dream-7B-Instruct.  All other settings are kept identical,
including the training data, LoRA configuration, optimizer, batch size, fixed
generation length, and the absence of VoidExpansion, matching the fixed-length
generation robustness experiments in Table~\ref{tab:blocksize-robustness-main}.
Under this matched setup,
VoidPadding achieves better accuracy than Early RainbowPadding on
both models.

  \section{Proofs for the \texttt{[EOS]} Overflow Analysis}
  \label{sec:proof}

  This appendix proves Theorem~\ref{thm:tail-eos-commit} and
  Theorem~\ref{thm:leftward-eos-spread}.  We use the notation and assumptions
  introduced in Section~\ref{sec:eos-analysis}: \(e\) denotes \texttt{[EOS]},
  \(F(m)=\Pr(\ell\leq m)\), and \(r=\min\{i:F(i-1)>1/2\}\).  For position \(i\),
  \(\bar{Y}_i\) is the training-time padded target token, \(p_i\) is the
  training-time token distribution with \(p_i(e)=\Pr(\bar{Y}_i=e)\), \(q_i\) is
  the inference-time model distribution at a masked canvas position, and
  \(q_i(e)\) is its \texttt{[EOS]} probability.  The argmax proposal is
  \(\hat{y}_i=\arg\max_v q_i(v)\), and \(Y_i\) is the committed token.  We write
  \(S_e=\{Y_r=\cdots=Y_L=e\}\) for the inference-time
  committed \texttt{[EOS]} suffix event and
  \(\bar{S}_e=\{\bar{Y}_r=\cdots=\bar{Y}_L=e\}\) for the corresponding
  training-time \texttt{[EOS]} padding suffix event.  A selected masked position is filled with its
  argmax proposal.
  We also use the elementary fact that if \(q_i(e)>1/2\), then no other token
  can have probability at least \(q_i(e)\).  Thus \(e\) is the unique argmax
  proposal at position \(i\), and if position \(i\) is selected, then \(Y_i=e\).

  \paragraph{Proof of Theorem~\ref{thm:tail-eos-commit}.}
  Consider any selected position \(j\in\{r,\dots,B\}\).  Since \(j\geq r\) and
  \(F\) is non-decreasing,
  \[
      F(j-1)\geq F(r-1)>1/2 .
  \]
  By the Stage 1 assumption in Section~\ref{sec:eos-analysis},
  \(q_j(e)=p_j(e)=F(j-1)\).  Hence \(q_j(e)>1/2\), so \(e\) is the unique
  argmax proposal at position \(j\).  Because \(j\) is selected, the committed
  token is
  \[
      Y_j=e .
  \]
  The argument applies to every selected \(j\in\{r,\dots,B\}\), proving the
  claim.  \hfill \(\square\)

  \paragraph{Proof of Theorem~\ref{thm:leftward-eos-spread}.}
  Assume \(S_e\) has occurred, and consider any selected position \(i<r\)
  satisfying \(2F(i-1)>F(r-1)\).  By the Stage 2 assumption,
  \[
      q_i(e\mid S_e)
      =
      \Pr(\bar{Y}_i=e\mid \bar{S}_e).
  \]
  For this theorem, we use the Stage 2 conditional assumption from
  Section~\ref{sec:eos-analysis}, together with its length-signal form for the
  suffix event \(\bar{S}_e\): for \(i<r\),
  \begin{align*}
      q_i(e\mid S_e)
      &=
      p_i(e\mid \bar{S}_e) \\
      &=
      \frac{F(i-1)}{F(r-1)} .
  \end{align*}
  The condition \(2F(i-1)>F(r-1)\) is exactly
  \[
      \frac{F(i-1)}{F(r-1)}>1/2 .
  \]
  Thus \(q_i(e\mid S_e)>1/2\), so \(e\) is the unique argmax proposal at position
  \(i\).  Since \(i\) is selected, the decoder commits
  \[
      Y_i=e .
  \]
  This proves the claim.  \hfill \(\square\)

\section{Dream Decoding Wrapper}
\label{app:dream-wrapper}

Algorithm~\ref{alg:dream_blockwise_commit} summarizes the blockwise commit
wrapper used for Dream evaluations.

  \begin{algorithm}[t]
  \caption{Blockwise Commit Wrapper for Dream Decoding}
  \label{alg:dream_blockwise_commit}
  \begin{algorithmic}[1]
  \Require Prompt $x_{1:p}$, generation length $L$, block size $B$, total denoising steps $T$
  \State Initialize $x_{p+1:p+L} \leftarrow [M]$
  \State Split the generation span into $K=L/B$ blocks $\{\mathcal{B}_k\}_{k=1}^K$; set $S=T/K$
  \For{$k=1$ to $K$}
      \For{$s=1$ to $S$}
          \State Run Dream on the full sequence $x$
          \State Obtain predictions $\hat{x}_i$ and negative entropy confidence scores $c_i$ for masked positions
          \State $\mathcal{E}_k \leftarrow \{i \in \mathcal{B}_k \mid x_i=[M]\}$
          \If{$\mathcal{E}_k=\emptyset$}
              \State \textbf{break}
          \EndIf
          \State $C_s \leftarrow \operatorname{TopK}_{i\in \mathcal{E}_k}(c_i)$
          \State $x_i \leftarrow \hat{x}_i,\quad \forall i\in C_s$
          \State Keep all positions $j\notin C_s$ unchanged
      \EndFor
  \EndFor
  \State \Return $x_{1:p+L}$
  \end{algorithmic}
  \end{algorithm}

Dream still predicts all masked positions globally at each denoising step, but the
wrapper restricts the TopK commit candidates to the masked positions inside the current
block.

\section{Fixed-Length Block-Size Sweeps on Instruction-Tuned Models}
\label{app:llada8b-instruct-blocksize-sweeps}

This section reports the full fixed-length block-size sweeps behind
Table~\ref{tab:blocksize-robustness-main}. The following LLaDA-8B-Instruct and Dream-7B-Instruct sweeps
follow the same setup: all runs use a fixed generation length of 512 without
VoidExpansion; we vary the block size over \(B\in\{32,64,128,256,512\}\) and
report the mean, best, and worst scores across the sweep. Mean is the arithmetic
average over the five block sizes, while Best and Worst are the maximum and
minimum scores over the five block sizes, respectively. Reasoning benchmarks
report accuracy, and code-generation benchmarks report pass@1.

\subsection{LLaDA-8B-Instruct Fixed-512 Block-Size Sweep}

For LLaDA-8B-Instruct and RainbowPadding experiments in the tables, rows with the \texttt{[EOS]} term.
label apply \texttt{[EOS]}Termination, while rows without this label follow the
standard VanillaStopping rule. For VoidPadding, the default row includes
\texttt{[EOS]}Termination, and the w/o \texttt{[EOS]} term. row reports the
ablation without this termination rule.

Table~\ref{tab:appendix-llada8b-instruct-blocksize-gsm8k} reports GSM8K
accuracy, Table~\ref{tab:appendix-llada8b-instruct-blocksize-math500} reports
MATH500 accuracy, Table~\ref{tab:appendix-llada8b-instruct-blocksize-humaneval}
reports HumanEval pass@1, and
Table~\ref{tab:appendix-llada8b-instruct-blocksize-mbpp} reports MBPP pass@1.

\subsection{Dream-7B-Instruct Fixed-512 Block-Length Sweep}

Figure~\ref{fig:dream512-instruct-block-sweep} reports the corresponding
Dream-7B-Instruct fixed-length block-length sweep under our block-wise Dream
decoding wrapper.

\section{\texttt{[EOS]}Termination Efficiency}
\label{app:fixed-length-efficiency}

\begin{figure}[t]
\centering
\scriptsize
\definecolor{fixedBlockNfe64}{HTML}{581C87}
\definecolor{fixedBlockNfe128}{HTML}{8B5CF6}
\definecolor{fixedBlockNfe512}{HTML}{DDD6FE}
\newcommand{\fixedblocknfebar}[4]{
  \pgfmathsetmacro{\h}{#3/120}
  \path[fill=#4, draw=none] (#1-0.085,0) rectangle (#1+0.085,\h);
  \node[font=\tiny, rotate=90, anchor=west, text=black] at (#1,\h+0.04) {#2};
}
\newcommand{\fixedblocknfegroup}[5]{
  \node[font=\scriptsize, anchor=north] at (#1+0.22,-0.26) {#2};
  #3#4#5
}
\resizebox{\columnwidth}{!}{
\begin{tikzpicture}[x=1cm,y=1cm]
  \draw[gray!45] (0,0) -- (7.20,0);
  \draw[gray!45] (0,0) -- (0,4.55);
  \foreach \y/\lab in {0/0,1/120,2/240,3/360,4/480} {
    \draw[gray!18] (0,\y) -- (7.20,\y);
    \node[font=\tiny, anchor=east, text=gray!70] at (-0.07,\y) {\lab};
  }
  \node[font=\scriptsize, rotate=90, anchor=center] at (-0.58,2.25) {NFE/example};

  \fixedblocknfegroup{0.42}{GSM8K}
    {\fixedblocknfebar{0.42}{127.9}{127.88}{fixedBlockNfe64}}
    {\fixedblocknfebar{0.62}{127.6}{127.59}{fixedBlockNfe128}}
    {\fixedblocknfebar{0.82}{128.3}{128.31}{fixedBlockNfe512}}

  \fixedblocknfegroup{1.82}{HumanEval}
    {\fixedblocknfebar{1.82}{102.8}{102.82}{fixedBlockNfe64}}
    {\fixedblocknfebar{2.02}{101.3}{101.34}{fixedBlockNfe128}}
    {\fixedblocknfebar{2.22}{96.3}{96.25}{fixedBlockNfe512}}

  \fixedblocknfegroup{3.22}{MATH500}
    {\fixedblocknfebar{3.22}{418.4}{418.43}{fixedBlockNfe64}}
    {\fixedblocknfebar{3.42}{422.4}{422.41}{fixedBlockNfe128}}
    {\fixedblocknfebar{3.62}{488.0}{488.00}{fixedBlockNfe512}}

  \fixedblocknfegroup{4.62}{MBPP}
    {\fixedblocknfebar{4.62}{69.2}{69.17}{fixedBlockNfe64}}
    {\fixedblocknfebar{4.82}{69.8}{69.79}{fixedBlockNfe128}}
    {\fixedblocknfebar{5.02}{69.4}{69.44}{fixedBlockNfe512}}

  \fixedblocknfegroup{6.02}{Avg.}
    {\fixedblocknfebar{6.02}{179.6}{179.58}{fixedBlockNfe64}}
    {\fixedblocknfebar{6.22}{180.3}{180.28}{fixedBlockNfe128}}
    {\fixedblocknfebar{6.42}{195.5}{195.50}{fixedBlockNfe512}}

  \node[font=\scriptsize, anchor=center] at (3.60,4.96) {
    \tikz{\shade[left color=fixedBlockNfe64, right color=fixedBlockNfe512, draw=none] (0,0) rectangle (0.28,0.10);}
    Block size \(B\): \textcolor{gray!65}{dark \(64\) \(\rightarrow\) light \(512\)}
  };
\end{tikzpicture}
}
\caption{LLaDA-8B-Instruct VoidPadding NFE/example for \(B\in\{64,128,512\}\). Avg. is arithmetic over four benchmarks.}
\label{fig:appendix-fixed-length-void-term-nfe-by-block}
\end{figure}
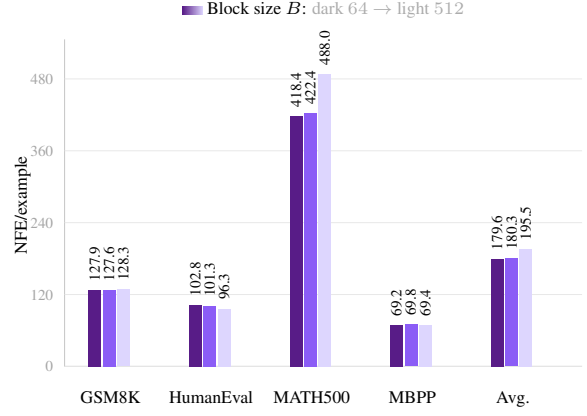

\begin{figure}[t]
\centering
\scriptsize
\definecolor{dreamFixedBlockNfe64}{HTML}{581C87}
\definecolor{dreamFixedBlockNfe128}{HTML}{8B5CF6}
\definecolor{dreamFixedBlockNfe512}{HTML}{DDD6FE}
\newcommand{\dreamfixedblocknfebar}[4]{
  \pgfmathsetmacro{\h}{#3/120}
  \path[fill=#4, draw=none] (#1-0.085,0) rectangle (#1+0.085,\h);
  \node[font=\tiny, rotate=90, anchor=west, text=black] at (#1,\h+0.04) {#2};
}
\newcommand{\dreamfixedblocknfegroup}[5]{
  \node[font=\scriptsize, anchor=north] at (#1+0.22,-0.26) {#2};
  #3#4#5
}
\resizebox{\columnwidth}{!}{
\begin{tikzpicture}[x=1cm,y=1cm]
  \draw[gray!45] (0,0) -- (7.20,0);
  \draw[gray!45] (0,0) -- (0,4.55);
  \foreach \y/\lab in {0/0,1/120,2/240,3/360,4/480} {
    \draw[gray!18] (0,\y) -- (7.20,\y);
    \node[font=\tiny, anchor=east, text=gray!70] at (-0.07,\y) {\lab};
  }
  \node[font=\scriptsize, rotate=90, anchor=center] at (-0.58,2.25) {NFE/example};

  \dreamfixedblocknfegroup{0.42}{GSM8K}
    {\dreamfixedblocknfebar{0.42}{150.2}{150.2}{dreamFixedBlockNfe64}}
    {\dreamfixedblocknfebar{0.62}{149.8}{149.8}{dreamFixedBlockNfe128}}
    {\dreamfixedblocknfebar{0.82}{172.0}{172.0}{dreamFixedBlockNfe512}}

  \dreamfixedblocknfegroup{1.82}{HumanEval}
    {\dreamfixedblocknfebar{1.82}{253.6}{253.6}{dreamFixedBlockNfe64}}
    {\dreamfixedblocknfebar{2.02}{254.8}{254.8}{dreamFixedBlockNfe128}}
    {\dreamfixedblocknfebar{2.22}{393.8}{393.8}{dreamFixedBlockNfe512}}

  \dreamfixedblocknfegroup{3.22}{MATH500}
    {\dreamfixedblocknfebar{3.22}{363.7}{363.7}{dreamFixedBlockNfe64}}
    {\dreamfixedblocknfebar{3.42}{366.4}{366.4}{dreamFixedBlockNfe128}}
    {\dreamfixedblocknfebar{3.62}{382.9}{382.9}{dreamFixedBlockNfe512}}

  \dreamfixedblocknfegroup{4.62}{MBPP}
    {\dreamfixedblocknfebar{4.62}{67.8}{67.8}{dreamFixedBlockNfe64}}
    {\dreamfixedblocknfebar{4.82}{69.1}{69.1}{dreamFixedBlockNfe128}}
    {\dreamfixedblocknfebar{5.02}{99.1}{99.1}{dreamFixedBlockNfe512}}

  \dreamfixedblocknfegroup{6.02}{Avg.}
    {\dreamfixedblocknfebar{6.02}{208.8}{208.825}{dreamFixedBlockNfe64}}
    {\dreamfixedblocknfebar{6.22}{210.0}{210.025}{dreamFixedBlockNfe128}}
    {\dreamfixedblocknfebar{6.42}{262.0}{261.950}{dreamFixedBlockNfe512}}

  \node[font=\scriptsize, anchor=center] at (3.60,4.96) {
    \tikz{\shade[left color=dreamFixedBlockNfe64, right color=dreamFixedBlockNfe512, draw=none] (0,0) rectangle (0.28,0.10);}
    Block size \(B\): \textcolor{gray!65}{dark \(64\) \(\rightarrow\) light \(512\)}
  };
\end{tikzpicture}
}
\caption{Dream-7B-Instruct VoidPadding NFE/example for \(B\in\{64,128,512\}\). Avg. is arithmetic over four benchmarks.}
\label{fig:appendix-dream-instruct-void-step7000-nfe-by-block}
\end{figure}
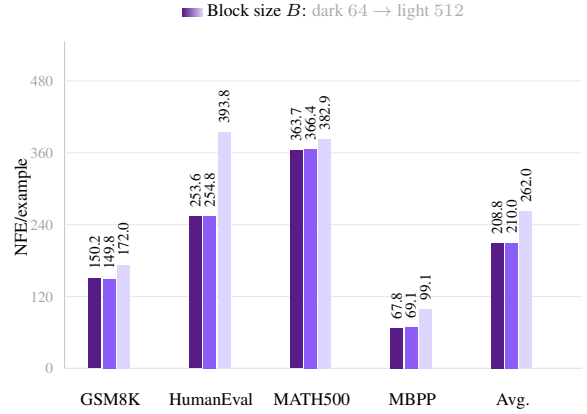

This appendix reports per-block fixed-length NFE/example values for the
Table~\ref{tab:blocksize-robustness-main} settings over \(B\in\{64,128,512\}\).
Figure~\ref{fig:appendix-fixed-length-void-term-nfe-by-block}
shows the LLaDA-8B-Instruct per-block NFE values, and
Figure~\ref{fig:appendix-dream-instruct-void-step7000-nfe-by-block} shows the
Dream-7B-Instruct per-block NFE values behind
Table~\ref{tab:fixed-length-efficiency-main}. Avg. entries are arithmetic
averages over benchmarks.

\let\fixedblocknfebar\relax
\let\fixedblocknfegroup\relax
\let\dreamfixedblocknfebar\relax
\let\dreamfixedblocknfegroup\relax

\section{Fixed-Length Block-Size Sweeps on Base Models}
\label{app:base-model-blocksize-sweeps}
The sweep configuration matches
Section~\ref{app:llada8b-instruct-blocksize-sweeps}: all runs use a fixed
generation length of 512 without VoidExpansion, and we vary the block size over
\(B\in\{32,64,128,256,512\}\). The only difference is that these diagnostics
start from base models and compare \texttt{[EOS]} Padding, RainbowPadding,
and VoidPadding variants.

\subsection{LLaDA-8B-Base Fixed-512 Block-Size Sweep}
\label{app:llada8b-base-blocksize-sweeps}

This subsection reports the LLaDA-8B-Base fixed-length block-size sweep. For
\texttt{[EOS]} Padding and RainbowPadding, rows with the \texttt{[EOS]} term.
label apply \texttt{[EOS]}Termination, while rows without this label follow the
standard VanillaStopping rule. For VoidPadding, the default row includes
\texttt{[EOS]}Termination, and the w/o \texttt{[EOS]} term. row reports the
ablation without this termination rule. GSM8K reports flexible exact match,
MATH500 reports accuracy, and HumanEval and MBPP report pass@1.
Table~\ref{tab:appendix-llada8b-base-blocksize-gsm8k} reports GSM8K,
Table~\ref{tab:appendix-llada8b-base-blocksize-math500} reports MATH500,
Table~\ref{tab:appendix-llada8b-base-blocksize-humaneval} reports HumanEval,
and Table~\ref{tab:appendix-llada8b-base-blocksize-mbpp} reports MBPP.

\subsection{Dream-7B-Base Fixed-512 Block-Length Sweep}

Figure~\ref{fig:dream512-base-block-sweep} reports the corresponding
Dream-7B-Base fixed-length block-length sweep under the same block-wise Dream
decoding wrapper.  

\section{Base-Model VoidExpansion Effectiveness and Hyperparameter Robustness}
\label{app:base-void-expansion-threshold-sweeps}

Unless otherwise stated, the base-model VoidExpansion sweeps use initial
length \(L_0=64\), tail-window size \(w=16\), block size \(B=32\), and
maximum length \(L_{\max}=512\).  The fixed-64 baseline denotes decoding from
a length-64 canvas with VoidExpansion disabled.

\subsection{LLaDA-8B-Base}

The common LLaDA-8B-Base VoidExpansion setting uses \(L_0=64\), \(w=16\),
\(B=32\), \(L_{\max}=512\), and
\((\tau_{\mathrm{nonvoid}},\tau_{\mathrm{gap}})=(0.09,-0.45)\).
Figure~\ref{fig:void-expansion-threshold-ablation} reports the corresponding
threshold sweep: it keeps \(L_0\), \(w\), \(B\), and \(L_{\max}\) fixed and
varies \(\tau_{\mathrm{nonvoid}}\) and \(\tau_{\mathrm{gap}}\).  Each panel
compares against the fixed-64 baseline and reports the benchmark mean as the
arithmetic average over GSM8K, MATH500, HumanEval, and MBPP.
Figure~\ref{fig:appendix-base-void-expansion-init-vs-instruct-fixed} reports the
initial-length ablation: it keeps \(w\), \(B\), \(L_{\max}\), and the threshold
pair fixed and varies \(L_0\) over \(64,128,192\), comparing fixed-length
decoding with VoidExpansion at the same lengths.  Across both sweeps, the
four-task mean changes by only about two points, and VoidExpansion remains
consistently stronger than the corresponding fixed-length baselines. 

\subsection{Dream-7B-Base}

The Dream-7B-Base threshold sweep is shown in
Figure~\ref{fig:dream-void-expansion-threshold-ablation-base}.  Its selected
setting is \((\tau_{\mathrm{nonvoid}},\tau_{\mathrm{gap}})=(0.12,0.65)\), and
the red ticks again mark the corresponding fixed-64 baseline, i.e., \(L_0=64\)
with VoidExpansion disabled.  As with LLaDA-8B-Base, the threshold sweep shows
only modest accuracy variation across settings, while VoidExpansion remains
consistently stronger than the fixed-64 baseline.

\section{VoidExpansion Hyperparameter Robustness}
\label{app:void-expansion-threshold-sweeps}

This appendix reports the complete VoidExpansion sweeps behind the
hyperparameter summary in Table~\ref{tab:void-expansion-hparam-summary}.  We
first present VoidPadding-finetuned LLaDA-8B-Instruct sweeps
over initial length \((L_0)\), tail-window size \(w\), block size \((B)\), and threshold pairs. 
We then
report the corresponding VoidPadding-finetuned Dream-7B-Instruct sweeps over threshold pairs.

\subsection{LLaDA-8B-Instruct}

This subsection provides the full LLaDA-8B-Instruct results summarized in
Section~\ref{sec:ablations}.  All runs use the
VoidPadding-finetuned LLaDA-8B-Instruct model with maximum generation length
\(L_{\max}=512\).  The default expansion setting, matching
Table~\ref{tab:gen-length-robustness}, uses initial length \(L_0=64\), tail
window \(w=16\), block size \(B=32\), and
\((\tau_{\mathrm{nonvoid}},\tau_{\mathrm{gap}})=(0.09,-0.45)\).

\paragraph{Initial-length robustness.}

Figure~\ref{fig:appendix-instruct-void-fixed-genlen} varies the initial length
\(L_0\in\{64,96,128,160\}\) while keeping the other expansion hyperparameters fixed, and compares
no-expansion fixed-length decoding against VoidExpansion at the same initial
lengths.  Figure~\ref{fig:appendix-void-initlen-nfe} reports the corresponding
average decode NFE.

\paragraph{Window-size robustness.}

Figure~\ref{fig:void-expansion-window-ablation} varies the tail-window size
\(w\in\{8,16,24,32\}\) while keeping the other settings fixed.
Figure~\ref{fig:appendix-void-window-nfe} reports the corresponding decode NFE
with the same benchmark grouping.  These figures provide the full data for the
tail-window row in
Table~\ref{tab:void-expansion-hparam-summary}.

\paragraph{Block-size robustness.}

Figure~\ref{fig:void-expansion-blocksize-ablation} varies the block size \(B\in\{16,32,64\}\)
while keeping the other settings fixed.
Figure~\ref{fig:appendix-void-block-nfe} reports the corresponding decode NFE
with the same benchmark grouping.  These figures provide the full data for the
block-size row in
Table~\ref{tab:void-expansion-hparam-summary}.

\paragraph{Threshold robustness.}

Figure~\ref{fig:void-expansion-threshold-ablation-instruct} varies
\(\tau_{\mathrm{nonvoid}}\) and \(\tau_{\mathrm{gap}}\) around the selected
setting from Table~\ref{tab:gen-length-robustness},
\((\tau_{\mathrm{nonvoid}},\tau_{\mathrm{gap}})=(0.09,-0.45)\), while keeping
\(L_0\), \(w\), \(B\), and \(L_{\max}\) fixed.
Figure~\ref{fig:appendix-void-threshold-nfe} reports the corresponding NFE
results.  These two figures provide the detailed threshold row behind
Table~\ref{tab:void-expansion-hparam-summary}.

\subsection{Dream-7B-Instruct}

Figure~\ref{fig:dream-void-expansion-threshold-ablation-instruct} reports the
VoidPadding-finetuned Dream-7B-Instruct + VoidExpansion threshold sweep under
our block-wise Dream decoding wrapper.  The selected setting from
Table~\ref{tab:dream-init64-expansion},
\((\tau_{\mathrm{nonvoid}},\tau_{\mathrm{gap}})=(0.15,0.55)\), is marked in
blue.  The red ticks show the fixed-64 baseline, i.e., decoding from a
length-64 canvas with VoidExpansion disabled, using the same
VoidPadding-finetuned checkpoint.  The scores remain stable across the
threshold grid, showing that Dream-7B-Instruct is also robust to the
VoidExpansion threshold choices.

\begin{figure}[t]
\centering
\scriptsize
\definecolor{effRho}{HTML}{D97706}
\definecolor{effDaedal}{HTML}{15803D}
\definecolor{effVoid}{HTML}{2563EB}
\begin{tikzpicture}
\begin{axis}[
    ybar,
    /pgf/bar width=5.8pt,
    width=\linewidth,
    height=0.72\linewidth,
    symbolic x coords={GSM8K,MATH500,HEval,MBPP},
    xtick=data,
    xticklabel style={font=\tiny},
    yticklabel style={font=\tiny},
    ylabel style={font=\tiny},
    ylabel={Total wall time (s)},
    ymin=0,
    ymax=50000,
    ymajorgrids=true,
    grid style={draw=gray!18},
    axis line style={gray!40},
    tick style={draw=gray!50},
    enlarge x limits=0.18,
    legend style={
      font=\scriptsize,
      draw=none,
      fill=none,
      at={(0.5,1.03)},
      anchor=south,
      legend columns=3,
      /tikz/every even column/.append style={column sep=1.0em}
    },
    legend image code/.code={
      \path[#1, draw=none] (0cm,-0.04cm) rectangle (0.24cm,0.06cm);
    },
]
\addplot+[fill=effRho, draw=none] coordinates
  {(GSM8K,20440.49) (MATH500,5874.55) (HEval,3547.86) (MBPP,27217.04)};
\addplot+[fill=effDaedal, draw=none] coordinates
  {(GSM8K,31846.30) (MATH500,17867.82) (HEval,4742.83) (MBPP,45945.10)};
\addplot+[fill=effVoid, draw=none] coordinates
  {(GSM8K,34469.40) (MATH500,21780.07) (HEval,847.75) (MBPP,9566.71)};
\legend{\(\rho\)-\texttt{[EOS]}, \textsc{Daedal}, VoidExpansion}
\end{axis}
\end{tikzpicture}
\caption{Total wall time on LLaDA-8B-Instruct under the init-64 setting. Bars compare \(\rho\)-\texttt{[EOS]}, \textsc{Daedal}, and VoidExpansion.}
\label{fig:appendix-expansion-efficiency-breakdown}
\end{figure}
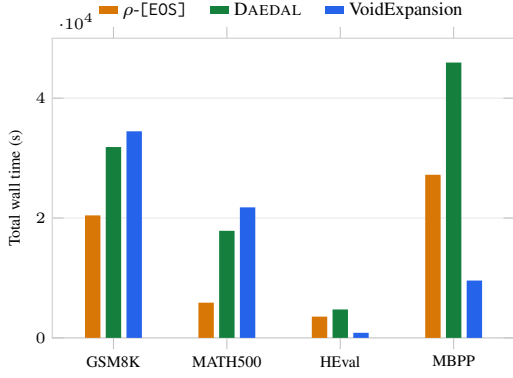
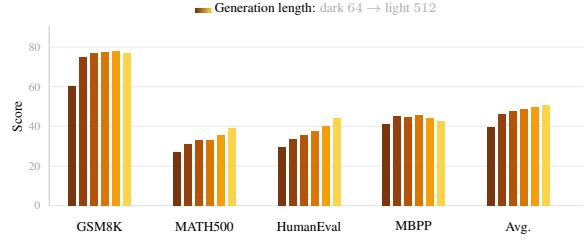
\begin{figure}[t]
\centering
\scriptsize
\definecolor{fixedLen64}{HTML}{78350F}
\definecolor{fixedLen128}{HTML}{92400E}
\definecolor{fixedLen160}{HTML}{B45309}
\definecolor{fixedLen192}{HTML}{D97706}
\definecolor{fixedLen256}{HTML}{F59E0B}
\definecolor{fixedLen512}{HTML}{FCD34D}
\newcommand{\standalonefixedlenbar}[4]{
  \pgfmathsetmacro{\h}{#3/20}
  \path[fill=#4, draw=none] (#1-0.07,0) rectangle (#1+0.07,\h);
}
\newcommand{\standalonefixedlengroup}[8]{
  \node[font=\scriptsize, anchor=north] at (#1+0.50,-0.28) {#2};
  #3#4#5#6#7#8
}
\resizebox{\columnwidth}{!}{
\begin{tikzpicture}[x=1cm,y=0.72cm]
  \draw[gray!45] (0,0) -- (9.70,0);
  \draw[gray!45] (0,0) -- (0,4.55);
  \foreach \y/\lab in {0/0,1/20,2/40,3/60,4/80} {
    \draw[gray!18] (0,\y) -- (9.70,\y);
    \node[font=\tiny, anchor=east, text=gray!70] at (-0.07,\y) {\lab};
  }
  \node[font=\scriptsize, rotate=90, anchor=center] at (-0.58,2.25) {Score};

  \standalonefixedlengroup{0.42}{GSM8K}
    {\standalonefixedlenbar{0.42}{60.2}{60.20}{fixedLen64}}
    {\standalonefixedlenbar{0.62}{74.7}{74.68}{fixedLen128}}
    {\standalonefixedlenbar{0.82}{76.9}{76.88}{fixedLen160}}
    {\standalonefixedlenbar{1.02}{77.6}{77.63}{fixedLen192}}
    {\standalonefixedlenbar{1.22}{78.0}{78.01}{fixedLen256}}
    {\standalonefixedlenbar{1.42}{76.7}{76.72}{fixedLen512}}

  \standalonefixedlengroup{2.32}{MATH500}
    {\standalonefixedlenbar{2.32}{26.8}{26.80}{fixedLen64}}
    {\standalonefixedlenbar{2.52}{30.8}{30.80}{fixedLen128}}
    {\standalonefixedlenbar{2.72}{33.2}{33.20}{fixedLen160}}
    {\standalonefixedlenbar{2.92}{33.0}{33.00}{fixedLen192}}
    {\standalonefixedlenbar{3.12}{35.8}{35.80}{fixedLen256}}
    {\standalonefixedlenbar{3.32}{39.0}{39.00}{fixedLen512}}

  \standalonefixedlengroup{4.22}{HumanEval}
    {\standalonefixedlenbar{4.22}{29.3}{29.27}{fixedLen64}}
    {\standalonefixedlenbar{4.42}{33.5}{33.54}{fixedLen128}}
    {\standalonefixedlenbar{4.62}{35.4}{35.37}{fixedLen160}}
    {\standalonefixedlenbar{4.82}{37.8}{37.80}{fixedLen192}}
    {\standalonefixedlenbar{5.02}{40.2}{40.24}{fixedLen256}}
    {\standalonefixedlenbar{5.22}{43.9}{43.90}{fixedLen512}}

  \standalonefixedlengroup{6.12}{MBPP}
    {\standalonefixedlenbar{6.12}{41.3}{41.27}{fixedLen64}}
    {\standalonefixedlenbar{6.32}{45.2}{45.17}{fixedLen128}}
    {\standalonefixedlenbar{6.52}{44.8}{44.76}{fixedLen160}}
    {\standalonefixedlenbar{6.72}{45.6}{45.59}{fixedLen192}}
    {\standalonefixedlenbar{6.92}{44.4}{44.35}{fixedLen256}}
    {\standalonefixedlenbar{7.12}{42.7}{42.71}{fixedLen512}}

  \standalonefixedlengroup{8.02}{Avg.}
    {\standalonefixedlenbar{8.02}{39.4}{39.38}{fixedLen64}}
    {\standalonefixedlenbar{8.22}{46.0}{46.05}{fixedLen128}}
    {\standalonefixedlenbar{8.42}{47.6}{47.55}{fixedLen160}}
    {\standalonefixedlenbar{8.62}{48.5}{48.51}{fixedLen192}}
    {\standalonefixedlenbar{8.82}{49.6}{49.60}{fixedLen256}}
    {\standalonefixedlenbar{9.02}{50.6}{50.58}{fixedLen512}}

  \node[font=\scriptsize, anchor=center] at (4.85,4.96) {
    \tikz{\shade[left color=fixedLen64, right color=fixedLen512, draw=none] (0,0) rectangle (0.28,0.10);}
    Generation length: \textcolor{gray!65}{dark \(64\) \(\rightarrow\) light \(512\)}
  };
\end{tikzpicture}
}
\caption{Fixed generation length sensitivity for LLaDA-8B-Instruct with block size \(B=32\) and generation length \(L\in\{64,128,160,192,256,512\}\). Darker orange denotes shorter \(L\); Avg. is the arithmetic average over benchmarks.}
\label{fig:instruct-genlen-sweep}
\end{figure}
\begin{figure}[t]
\centering
\scriptsize
\definecolor{accWindow8}{HTML}{1E3A8A}
\definecolor{accWindow16}{HTML}{1D4ED8}
\definecolor{accWindow24}{HTML}{60A5FA}
\definecolor{accWindow32}{HTML}{BFDBFE}
\newcommand{\windowaccbar}[4]{
  \pgfmathsetmacro{\h}{#3/20}
  \path[fill=#4, draw=none] (#1-0.085,0) rectangle (#1+0.085,\h);
  \node[font=\tiny, rotate=90, anchor=west, text=black] at (#1,\h+0.04) {#2};
}
\newcommand{\windowaccgroup}[6]{
  \node[font=\scriptsize, anchor=north] at (#1+0.33,-0.26) {#2};
  #3#4#5#6
}
\resizebox{\columnwidth}{!}{
\begin{tikzpicture}[x=1cm,y=1cm]
  \draw[gray!45] (0,0) -- (7.75,0);
  \draw[gray!45] (0,0) -- (0,4.55);
  \foreach \y/\lab in {0/0,1/20,2/40,3/60,4/80} {
    \draw[gray!18] (0,\y) -- (7.75,\y);
    \node[font=\tiny, anchor=east, text=gray!70] at (-0.07,\y) {\lab};
  }
  \node[font=\scriptsize, rotate=90, anchor=center] at (-0.58,2.25) {Score};

  \windowaccgroup{0.42}{GSM8K}
    {\windowaccbar{0.42}{79.4}{79.38}{accWindow8}}
    {\windowaccbar{0.62}{78.3}{78.32}{accWindow16}}
    {\windowaccbar{0.82}{78.5}{78.54}{accWindow24}}
    {\windowaccbar{1.02}{78.3}{78.32}{accWindow32}}

  \windowaccgroup{1.95}{MATH500}
    {\windowaccbar{1.95}{39.6}{39.60}{accWindow8}}
    {\windowaccbar{2.15}{39.6}{39.60}{accWindow16}}
    {\windowaccbar{2.35}{39.8}{39.80}{accWindow24}}
    {\windowaccbar{2.55}{39.8}{39.80}{accWindow32}}

  \windowaccgroup{3.48}{HumanEval}
    {\windowaccbar{3.48}{42.7}{42.68}{accWindow8}}
    {\windowaccbar{3.68}{42.7}{42.68}{accWindow16}}
    {\windowaccbar{3.88}{42.7}{42.68}{accWindow24}}
    {\windowaccbar{4.08}{41.5}{41.46}{accWindow32}}

  \windowaccgroup{5.01}{MBPP}
    {\windowaccbar{5.01}{45.6}{45.59}{accWindow8}}
    {\windowaccbar{5.21}{46.5}{46.51}{accWindow16}}
    {\windowaccbar{5.41}{46.2}{46.20}{accWindow24}}
    {\windowaccbar{5.61}{46.0}{46.00}{accWindow32}}

  \windowaccgroup{6.54}{Mean}
    {\windowaccbar{6.54}{51.8}{51.81}{accWindow8}}
    {\windowaccbar{6.74}{51.8}{51.78}{accWindow16}}
    {\windowaccbar{6.94}{51.8}{51.81}{accWindow24}}
    {\windowaccbar{7.14}{51.4}{51.40}{accWindow32}}

  \node[font=\scriptsize, anchor=center] at (3.90,4.96) {
    \tikz{\shade[left color=accWindow8, right color=accWindow32, draw=none] (0,0) rectangle (0.28,0.10);}
    Tail window \(w\): \textcolor{gray!65}{dark \(8\) \(\rightarrow\) light \(32\)}
  };
\end{tikzpicture}
}
\caption{Window-size accuracy statistics for VoidPadding-finetuned LLaDA-8B-Instruct + VoidExpansion. Bars are grouped by benchmark; within each group, the four bars vary the tail window \(w\in\{8,16,24,32\}\). Darker blue denotes smaller \(w\). Labels give the exact score. GSM8K and MATH500 report accuracy, HumanEval and MBPP report pass@1, and Mean is the arithmetic average over the four benchmarks.}
\label{fig:void-expansion-window-ablation}
\end{figure}
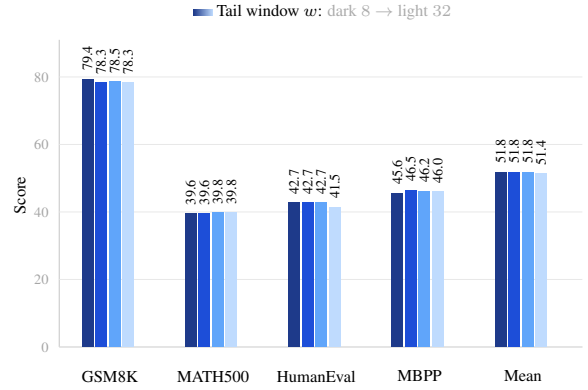
\begin{figure}[t]
\centering
\scriptsize
\definecolor{window8}{HTML}{581C87}
\definecolor{window16}{HTML}{7E22CE}
\definecolor{window24}{HTML}{A855F7}
\definecolor{window32}{HTML}{DDD6FE}
\newcommand{\windownfebar}[4]{
  \pgfmathsetmacro{\h}{#3/100}
  \path[fill=#4, draw=none] (#1-0.085,0) rectangle (#1+0.085,\h);
  \node[font=\tiny, rotate=90, anchor=west, text=black] at (#1,\h+0.04) {#2};
}
\newcommand{\windownfegroup}[6]{
  \node[font=\scriptsize, anchor=north] at (#1+0.33,-0.26) {#2};
  #3#4#5#6
}
\resizebox{\columnwidth}{!}{
\begin{tikzpicture}[x=1cm,y=1cm]
  \draw[gray!45] (0,0) -- (7.75,0);
  \draw[gray!45] (0,0) -- (0,4.55);
  \foreach \y/\lab in {0/0,1/100,2/200,3/300,4/400} {
    \draw[gray!18] (0,\y) -- (7.75,\y);
    \node[font=\tiny, anchor=east, text=gray!70] at (-0.07,\y) {\lab};
  }
  \node[font=\scriptsize, rotate=90, anchor=center] at (-0.58,2.25) {NFE/example};

  \windownfegroup{0.42}{GSM8K}
    {\windownfebar{0.42}{130.6}{130.55}{window8}}
    {\windownfebar{0.62}{133.8}{133.78}{window16}}
    {\windownfebar{0.82}{133.4}{133.35}{window24}}
    {\windownfebar{1.02}{135.1}{135.10}{window32}}

  \windownfegroup{1.95}{MATH500}
    {\windownfebar{1.95}{418.1}{418.05}{window8}}
    {\windownfebar{2.15}{418.1}{418.05}{window16}}
    {\windownfebar{2.35}{415.2}{415.16}{window24}}
    {\windownfebar{2.55}{410.3}{410.30}{window32}}

  \windownfegroup{3.48}{HumanEval}
    {\windownfebar{3.48}{80.3}{80.30}{window8}}
    {\windownfebar{3.68}{73.1}{73.06}{window16}}
    {\windownfebar{3.88}{68.8}{68.84}{window24}}
    {\windownfebar{4.08}{65.3}{65.27}{window32}}

  \windownfegroup{5.01}{MBPP}
    {\windownfebar{5.01}{63.7}{63.70}{window8}}
    {\windownfebar{5.21}{63.5}{63.50}{window16}}
    {\windownfebar{5.41}{63.2}{63.21}{window24}}
    {\windownfebar{5.61}{63.9}{63.91}{window32}}

  \windownfegroup{6.54}{Avg.}
    {\windownfebar{6.54}{173.2}{173.15}{window8}}
    {\windownfebar{6.74}{172.1}{172.10}{window16}}
    {\windownfebar{6.94}{170.1}{170.14}{window24}}
    {\windownfebar{7.14}{168.7}{168.65}{window32}}

  \node[font=\scriptsize, anchor=center] at (3.90,4.96) {
    \tikz{\shade[left color=window8, right color=window32, draw=none] (0,0) rectangle (0.28,0.10);}
    Tail window \(w\): \textcolor{gray!65}{dark \(8\) \(\rightarrow\) light \(32\)}
  };
\end{tikzpicture}
}
\caption{Window-size NFE statistics for VoidPadding-finetuned LLaDA-8B-Instruct + VoidExpansion. Bars are grouped by benchmark; within each group, the four bars vary the tail window \(w\in\{8,16,24,32\}\). Darker purple denotes smaller \(w\). Labels report average decode NFE. Avg. is the arithmetic average over the four benchmarks.}
\label{fig:appendix-void-window-nfe}
\end{figure}
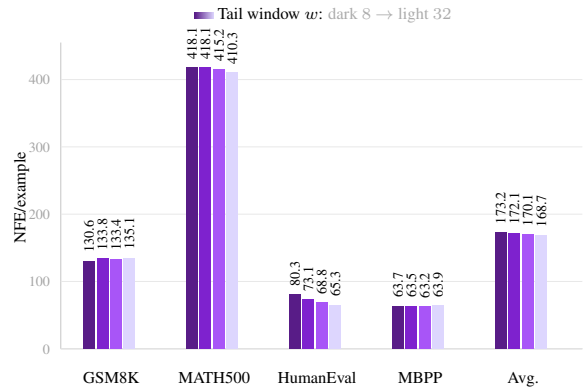

\begin{table*}[t]
\centering
\scriptsize
\setlength{\tabcolsep}{4pt}
\resizebox{\textwidth}{!}{
\begin{tabular}{lcccccccc}
\toprule
\multirow{2}{*}{Method} &
\multicolumn{5}{c}{Block size \(B\)} &
\multicolumn{3}{c}{Summary} \\
\cmidrule(lr){2-6}\cmidrule(lr){7-9}
& 32 & 64 & 128 & 256 & 512 & Mean & Best & Worst \\
\midrule
LLaDA-8B-Instruct & 76.72 & \underline{76.80} & 70.96 & 44.96 & 28.43 & 59.57 & 76.80 & 28.43 \\
LLaDA-8B-Instruct + \texttt{[EOS]} term. & \textbf{77.71} & 76.65 & 69.60 & 39.95 & 28.13 & 58.41 & \underline{77.71} & 28.13 \\
RainbowPadding & 73.77 & 74.53 & \underline{73.39} & \underline{72.25} & \underline{66.57} & \underline{72.10} & 74.53 & \underline{66.57} \\
RainbowPadding + \texttt{[EOS]} term. & 73.77 & 74.53 & \underline{73.39} & \underline{72.25} & \underline{66.57} & \underline{72.10} & 74.53 & \underline{66.57} \\
VoidPadding w/o \texttt{[EOS]} term. & 57.32 & 57.16 & 57.24 & 55.95 & 65.66 & 58.67 & 65.66 & 55.95 \\
VoidPadding w/o VOID ban & 75.06 & 64.67 & 14.71 & 11.14 & 14.40 & 36.00 & 75.06 & 11.14 \\
VoidPadding & \underline{77.41} & \textbf{77.79} & \textbf{78.39} & \textbf{78.24} & \textbf{78.62} & \textbf{78.09} & \textbf{78.62} & \textbf{77.41} \\
\bottomrule
\end{tabular}
}
\caption{GSM8K block-size sweep for LLaDA-8B-Instruct; cells report accuracy (\%).}
\label{tab:appendix-llada8b-instruct-blocksize-gsm8k}
\end{table*}

\begin{table*}[t]
\centering
\scriptsize
\setlength{\tabcolsep}{4pt}
\resizebox{\textwidth}{!}{
\begin{tabular}{lcccccccc}
\toprule
\multirow{2}{*}{Method} &
\multicolumn{5}{c}{Block size \(B\)} &
\multicolumn{3}{c}{Summary} \\
\cmidrule(lr){2-6}\cmidrule(lr){7-9}
& 32 & 64 & 128 & 256 & 512 & Mean & Best & Worst \\
\midrule
LLaDA-8B-Instruct & \underline{39.00} & 39.60 & 37.60 & 30.60 & 16.60 & 32.68 & 39.60 & 16.60 \\
LLaDA-8B-Instruct + \texttt{[EOS]} term. & \underline{39.00} & 39.60 & 36.80 & 30.40 & 16.60 & 32.48 & 39.60 & 16.60 \\
RainbowPadding & 33.20 & 33.20 & 34.00 & 34.00 & \textbf{27.80} & 32.44 & 34.00 & \textbf{27.80} \\
RainbowPadding + \texttt{[EOS]} term. & 33.20 & 33.20 & 34.00 & 34.00 & \textbf{27.80} & 32.44 & 34.00 & \textbf{27.80} \\
VoidPadding w/o \texttt{[EOS]} term. & \textbf{39.60} & \textbf{40.60} & \textbf{41.60} & \textbf{40.20} & \underline{27.60} & \textbf{37.92} & \textbf{41.60} & \underline{27.60} \\
VoidPadding w/o VOID ban & \textbf{39.60} & \underline{39.80} & 38.20 & 32.20 & 14.20 & 32.80 & \underline{39.80} & 14.20 \\
VoidPadding & \textbf{39.60} & \textbf{40.60} & \textbf{41.60} & \underline{40.00} & \underline{27.60} & \underline{37.88} & \textbf{41.60} & \underline{27.60} \\
\bottomrule
\end{tabular}
}
\caption{MATH500 block-size sweep for LLaDA-8B-Instruct; cells report accuracy (\%).}
\label{tab:appendix-llada8b-instruct-blocksize-math500}
\end{table*}

\begin{table*}[t]
\centering
\scriptsize
\setlength{\tabcolsep}{4pt}
\resizebox{\textwidth}{!}{
\begin{tabular}{lcccccccc}
\toprule
\multirow{2}{*}{Method} &
\multicolumn{5}{c}{Block size \(B\)} &
\multicolumn{3}{c}{Summary} \\
\cmidrule(lr){2-6}\cmidrule(lr){7-9}
& 32 & 64 & 128 & 256 & 512 & Mean & Best & Worst \\
\midrule
\multicolumn{9}{l}{\textit{pass@1}} \\
LLaDA-8B-Instruct & \underline{43.90} & \underline{42.68} & \textbf{43.90} & \underline{42.07} & \textbf{44.51} & \textbf{43.41} & \underline{44.51} & \textbf{42.07} \\
LLaDA-8B-Instruct + \texttt{[EOS]} term. & 42.68 & \underline{42.68} & \underline{43.29} & 40.24 & \underline{43.90} & \underline{42.56} & 43.90 & 40.24 \\
RainbowPadding & \textbf{45.12} & \textbf{43.90} & \underline{43.29} & \textbf{43.29} & 35.98 & 42.32 & \textbf{45.12} & 35.98 \\
RainbowPadding + \texttt{[EOS]} term. & \textbf{45.12} & \textbf{43.90} & \underline{43.29} & \textbf{43.29} & 35.98 & 42.32 & \textbf{45.12} & 35.98 \\
VoidPadding w/o \texttt{[EOS]} term. & 19.51 & 20.73 & 22.56 & 23.78 & 20.73 & 21.46 & 23.78 & 19.51 \\
VoidPadding w/o VOID ban & 29.88 & 17.07 & 10.37 & 6.10 & 6.10 & 13.90 & 29.88 & 6.10 \\
VoidPadding & 42.07 & 41.46 & 42.07 & \underline{42.07} & 41.46 & 41.83 & 42.07 & \underline{41.46} \\
\bottomrule
\end{tabular}
}
\caption{HumanEval block-size sweep for LLaDA-8B-Instruct, with pass@1 (\%).}
\label{tab:appendix-llada8b-instruct-blocksize-humaneval}
\end{table*}

\begin{table*}[t]
\centering
\scriptsize
\setlength{\tabcolsep}{4pt}
\resizebox{\textwidth}{!}{
\begin{tabular}{lcccccccc}
\toprule
\multirow{2}{*}{Method} &
\multicolumn{5}{c}{Block size \(B\)} &
\multicolumn{3}{c}{Summary} \\
\cmidrule(lr){2-6}\cmidrule(lr){7-9}
& 32 & 64 & 128 & 256 & 512 & Mean & Best & Worst \\
\midrule
\multicolumn{9}{l}{\textit{pass@1}} \\
LLaDA-8B-Instruct & 42.81 & 41.48 & 39.12 & 33.98 & 37.68 & 39.01 & 42.81 & 33.98 \\
LLaDA-8B-Instruct + \texttt{[EOS]} term. & \underline{43.12} & 42.09 & 39.73 & 35.22 & 38.71 & 39.77 & \underline{43.12} & 35.22 \\
RainbowPadding & 42.51 & \underline{42.71} & \underline{43.02} & \underline{43.12} & \underline{42.51} & \underline{42.77} & \underline{43.12} & \underline{42.51} \\
RainbowPadding + \texttt{[EOS]} term. & 42.40 & 42.61 & \underline{43.02} & \underline{43.12} & \underline{42.51} & 42.73 & \underline{43.12} & 42.40 \\
VoidPadding w/o \texttt{[EOS]} term. & \textbf{44.87} & \textbf{44.66} & \textbf{44.87} & \textbf{44.87} & \textbf{44.97} & \textbf{44.85} & \textbf{44.97} & \textbf{44.66} \\
VoidPadding w/o VOID ban & 41.27 & 25.36 & 16.94 & 17.66 & 17.56 & 23.76 & 41.27 & 16.94 \\
VoidPadding & \textbf{44.87} & \textbf{44.66} & \textbf{44.87} & \textbf{44.87} & \textbf{44.97} & \textbf{44.85} & \textbf{44.97} & \textbf{44.66} \\
\bottomrule
\end{tabular}
}
\caption{MBPP block-size sweep for LLaDA-8B-Instruct, with pass@1 (\%).}
\label{tab:appendix-llada8b-instruct-blocksize-mbpp}
\end{table*}

\begin{table*}[t]
\centering
\scriptsize
\setlength{\tabcolsep}{4pt}
\resizebox{\textwidth}{!}{
\begin{tabular}{lcccccccc}
\toprule
\multirow{2}{*}{Method} &
\multicolumn{5}{c}{Block size \(B\)} &
\multicolumn{3}{c}{Summary} \\
\cmidrule(lr){2-6}\cmidrule(lr){7-9}
& 32 & 64 & 128 & 256 & 512 & Mean & Best & Worst \\
\midrule
LLaDA-8B-Base + \texttt{[EOS]} Padding & \textbf{77.41} & 58.76 & 15.31 & 7.88 & 11.83 & 34.24 & \textbf{77.41} & 7.88 \\
  LLaDA-8B-Base + \texttt{[EOS]} Padding + \texttt{[EOS]} term.
  & \textbf{77.41} & 58.76 & 15.31 & 7.88 & 11.83
  & 34.24 & \textbf{77.41} & 7.88 \\
RainbowPadding & \underline{76.27} & \underline{76.12} & \textbf{76.12} & \textbf{75.74} & \textbf{75.51} & \textbf{75.95} & 76.27 & \textbf{75.51} \\
RainbowPadding + \texttt{[EOS]} term. & 75.51 & \underline{76.12} & 75.66 & \underline{75.59} & \textbf{75.51} & 75.68 & 76.12 & \textbf{75.51} \\
VoidPadding \texttt{w/o [EOS] term.} & 3.11 & 3.03 & 3.41 & 3.26 & 2.88 & 3.14 & 3.41 & 2.88 \\
VoidPadding w/o VOID ban & 75.44 & 67.32 & 46.85 & 42.76 & 43.14 & 55.10 & 75.44 & 42.76 \\
VoidPadding & 75.51 & \textbf{76.57} & \underline{75.82} & 75.28 & \underline{75.28} & \underline{75.69} & \underline{76.57} & \underline{75.28} \\
\bottomrule
\end{tabular}
}
\caption{GSM8K block-size sweep for LLaDA-8B-Base after padding LoRA SFT; cells report accuracy (\%).}
\label{tab:appendix-llada8b-base-blocksize-gsm8k}
\end{table*}

\begin{table*}[t]
\centering
\scriptsize
\setlength{\tabcolsep}{4pt}
\resizebox{\textwidth}{!}{
\begin{tabular}{lcccccccc}
\toprule
\multirow{2}{*}{Method} &
\multicolumn{5}{c}{Block size \(B\)} &
\multicolumn{3}{c}{Summary} \\
\cmidrule(lr){2-6}\cmidrule(lr){7-9}
& 32 & 64 & 128 & 256 & 512 & Mean & Best & Worst \\
\midrule
LLaDA-8B-Base + \texttt{[EOS]} Padding & 37.80 & 36.80 & 34.00 & 29.40 & 20.60 & 31.72 & 37.80 & 20.60 \\
LLaDA-8B-Base + \texttt{[EOS]} Padding + \texttt{[EOS]} term. & 37.80 & 36.80 & 34.00 & 29.40 & 20.60 & 31.72 & 37.80 & 20.60 \\
RainbowPadding & 34.00 & 33.40 & 34.80 & 35.00 & \underline{34.80} & 34.40 & 35.00 & 33.40 \\
RainbowPadding + \texttt{[EOS]} term. & 35.80 & 35.80 & \underline{36.00} & \underline{36.00} & \textbf{36.60} & \underline{36.04} & 36.60 & \textbf{35.80} \\
VoidPadding \texttt{w/o [EOS] term.} & \textbf{38.40} & \textbf{40.00} & \textbf{39.20} & \textbf{37.40} & \underline{34.00} & \textbf{37.80} & \textbf{40.00} & \underline{34.00} \\
VoidPadding w/o VOID ban & \underline{38.20} & \underline{39.20} & 31.00 & 23.00 & 23.00 & 30.88 & \underline{39.20} & 23.00 \\
VoidPadding & \textbf{38.40} & \textbf{40.00} & \textbf{39.20} & \textbf{37.40} & \underline{34.00} & \textbf{37.80} & \textbf{40.00} & \underline{34.00} \\
\bottomrule
\end{tabular}
}
\caption{MATH500 block-size sweep for LLaDA-8B-Base after padding LoRA SFT; cells report accuracy (\%).}
\label{tab:appendix-llada8b-base-blocksize-math500}
\end{table*}

\begin{table*}[t]
\centering
\scriptsize
\setlength{\tabcolsep}{4pt}
\resizebox{\textwidth}{!}{
\begin{tabular}{lcccccccc}
\toprule
\multirow{2}{*}{Method} &
\multicolumn{5}{c}{Block size \(B\)} &
\multicolumn{3}{c}{Summary} \\
\cmidrule(lr){2-6}\cmidrule(lr){7-9}
& 32 & 64 & 128 & 256 & 512 & Mean & Best & Worst \\
\midrule
\multicolumn{9}{l}{\textit{pass@1}} \\
LLaDA-8B-Base + \texttt{[EOS]} Padding & 37.20 & 26.83 & 15.85 & 13.41 & 12.20 & 21.10 & 37.20 & 12.20 \\
LLaDA-8B-Base + \texttt{[EOS]} Padding + \texttt{[EOS]} term. & 37.20 & 26.83 & 15.85 & 13.41 & 12.20 & 21.10 & 37.20 & 12.20 \\
RainbowPadding & \textbf{38.41} & \textbf{39.02} & \textbf{38.41} & \textbf{38.41} & \textbf{38.41} & \textbf{38.53} & \textbf{39.02} & \textbf{38.41} \\
RainbowPadding + \texttt{[EOS]} term. & \underline{37.80} & \underline{38.41} & \textbf{38.41} & \textbf{38.41} & \textbf{38.41} & \underline{38.29} & \underline{38.41} & \underline{37.80} \\
VoidPadding \texttt{w/o [EOS] term.} & 14.63 & 17.07 & 22.56 & \underline{23.17} & 8.54 & 17.19 & 23.17 & 8.54 \\
VoidPadding w/o VOID ban & 35.37 & 18.29 & 12.80 & 14.63 & \underline{15.24} & 19.27 & 35.37 & 12.80 \\
VoidPadding & \underline{37.80} & 37.20 & \underline{37.80} & \textbf{38.41} & \textbf{38.41} & 37.92 & \underline{38.41} & 37.20 \\
\bottomrule
\end{tabular}
}
\caption{HumanEval block-size sweep for LLaDA-8B-Base after padding LoRA SFT, with pass@1 (\%).}
\label{tab:appendix-llada8b-base-blocksize-humaneval}
\end{table*}

\begin{table*}[t]
\centering
\scriptsize
\setlength{\tabcolsep}{4pt}
\resizebox{\textwidth}{!}{
\begin{tabular}{lcccccccc}
\toprule
\multirow{2}{*}{Method} &
\multicolumn{5}{c}{Block size \(B\)} &
\multicolumn{3}{c}{Summary} \\
\cmidrule(lr){2-6}\cmidrule(lr){7-9}
& 32 & 64 & 128 & 256 & 512 & Mean & Best & Worst \\
\midrule
\multicolumn{9}{l}{\textit{pass@1}} \\
LLaDA-8B-Base + \texttt{[EOS]} Padding & 43.22 & 32.55 & 23.72 & 23.82 & 22.18 & 29.10 & 43.22 & 22.18 \\
LLaDA-8B-Base + \texttt{[EOS]} Padding + \texttt{[EOS]} term. & 43.22 & 32.55 & 23.72 & 23.82 & 22.18 & 29.10 & 43.22 & 22.18 \\
RainbowPadding & 43.33 & \underline{43.53} & \underline{43.74} & \underline{43.53} & \underline{43.94} & \underline{43.61} & \underline{43.94} & \underline{43.33} \\
RainbowPadding + \texttt{[EOS]} term. & 43.33 & \underline{43.53} & \underline{43.74} & \underline{43.53} & \underline{43.94} & \underline{43.61} & \underline{43.94} & \underline{43.33} \\
VoidPadding \texttt{w/o [EOS] term.} & \textbf{44.35} & \textbf{44.87} & \textbf{44.66} & \textbf{44.76} & \textbf{44.76} & \textbf{44.68} & \textbf{44.87} & \textbf{44.35} \\
VoidPadding w/o VOID ban & \underline{43.63} & 29.98 & 21.56 & 20.64 & 18.38 & 26.84 & 43.63 & 18.38 \\
VoidPadding & \textbf{44.35} & \textbf{44.87} & \textbf{44.66} & \textbf{44.76} & \textbf{44.76} & \textbf{44.68} & \textbf{44.87} & \textbf{44.35} \\
\bottomrule
\end{tabular}
}
\caption{MBPP block-size sweep for LLaDA-8B-Base after padding LoRA SFT, with pass@1 (\%).}
\label{tab:appendix-llada8b-base-blocksize-mbpp}
\end{table*}

\clearpage
\input{figures/dream512_instruct_block_sweep}
\clearpage
\input{tables/appendix_instruct_void_fixed_genlen}
\clearpage

\begin{figure*}[t]
\centering
\begin{subfigure}[t]{0.19\textwidth}
\centering
\resizebox{\linewidth}{!}{
\begin{tikzpicture}
\VoidTauInstructHeatCell{0}{3}{78.62}{52}{79}
\VoidTauInstructHeatCell{1}{3}{78.62}{52}{79}
\VoidTauInstructHeatCell{2}{3}{78.32}{52}{79}
\VoidTauInstructHeatCell{3}{3}{77.79}{52}{79}
\VoidTauInstructHeatCell{0}{2}{78.62}{52}{79}
\VoidTauInstructHeatCell{1}{2}{78.62}{52}{79}
\VoidTauInstructHeatCell{2}{2}{78.32}{52}{79}
\VoidTauInstructHeatCell{3}{2}{77.79}{52}{79}
\VoidTauInstructHeatCell{0}{1}{78.62}{52}{79}
\VoidTauInstructHeatCell{1}{1}{78.62}{52}{79}
\VoidTauInstructHeatCell{2}{1}{78.32}{52}{79}
\VoidTauInstructHeatCell{3}{1}{77.79}{52}{79}
\VoidTauInstructHeatCell{0}{0}{78.62}{52}{79}
\VoidTauInstructHeatCell{1}{0}{78.62}{52}{79}
\VoidTauInstructHeatCell{2}{0}{78.32}{52}{79}
\VoidTauInstructHeatCell{3}{0}{77.79}{52}{79}
\VoidTauInstructPanelFrame{GSM8K}
\VoidTauInstructColorScale{52}{79}
\VoidTauInstructBaselineMark{52.84}{fixlen=64}{52}{79}{red!95!black}
\end{tikzpicture}
}
\end{subfigure}
\hfill
\begin{subfigure}[t]{0.19\textwidth}
\centering
\resizebox{\linewidth}{!}{
\begin{tikzpicture}
\VoidTauInstructHeatCell{0}{3}{39.60}{26}{40}
\VoidTauInstructHeatCell{1}{3}{39.60}{26}{40}
\VoidTauInstructHeatCell{2}{3}{39.60}{26}{40}
\VoidTauInstructHeatCell{3}{3}{39.80}{26}{40}
\VoidTauInstructHeatCell{0}{2}{39.60}{26}{40}
\VoidTauInstructHeatCell{1}{2}{39.60}{26}{40}
\VoidTauInstructHeatCell{2}{2}{39.60}{26}{40}
\VoidTauInstructHeatCell{3}{2}{39.80}{26}{40}
\VoidTauInstructHeatCell{0}{1}{39.60}{26}{40}
\VoidTauInstructHeatCell{1}{1}{39.60}{26}{40}
\VoidTauInstructHeatCell{2}{1}{39.60}{26}{40}
\VoidTauInstructHeatCell{3}{1}{39.80}{26}{40}
\VoidTauInstructHeatCell{0}{0}{39.60}{26}{40}
\VoidTauInstructHeatCell{1}{0}{39.60}{26}{40}
\VoidTauInstructHeatCell{2}{0}{39.60}{26}{40}
\VoidTauInstructHeatCell{3}{0}{39.80}{26}{40}
\VoidTauInstructPanelFrame{MATH500}
\VoidTauInstructColorScale{26}{40}
\VoidTauInstructBaselineMark{26.20}{fixlen=64}{26}{40}{red!95!black}
\end{tikzpicture}
}
\end{subfigure}
\hfill
\begin{subfigure}[t]{0.19\textwidth}
\centering
\resizebox{\linewidth}{!}{
\begin{tikzpicture}
\VoidTauInstructHeatCell{0}{3}{42.68}{37}{43}
\VoidTauInstructHeatCell{1}{3}{40.85}{37}{43}
\VoidTauInstructHeatCell{2}{3}{42.68}{37}{43}
\VoidTauInstructHeatCell{3}{3}{40.85}{37}{43}
\VoidTauInstructHeatCell{0}{2}{42.68}{37}{43}
\VoidTauInstructHeatCell{1}{2}{40.85}{37}{43}
\VoidTauInstructHeatCell{2}{2}{42.68}{37}{43}
\VoidTauInstructHeatCell{3}{2}{40.85}{37}{43}
\VoidTauInstructHeatCell{0}{1}{42.68}{37}{43}
\VoidTauInstructHeatCell{1}{1}{40.85}{37}{43}
\VoidTauInstructHeatCell{2}{1}{42.68}{37}{43}
\VoidTauInstructHeatCell{3}{1}{40.85}{37}{43}
\VoidTauInstructHeatCell{0}{0}{42.68}{37}{43}
\VoidTauInstructHeatCell{1}{0}{40.85}{37}{43}
\VoidTauInstructHeatCell{2}{0}{42.68}{37}{43}
\VoidTauInstructHeatCell{3}{0}{40.85}{37}{43}
\VoidTauInstructPanelFrame{HumanEval}
\VoidTauInstructColorScale{37}{43}
\VoidTauInstructBaselineMark{37.20}{fixlen=64}{37}{43}{red!95!black}
\end{tikzpicture}
}
\end{subfigure}
\hfill
\begin{subfigure}[t]{0.19\textwidth}
\centering
\resizebox{\linewidth}{!}{
\begin{tikzpicture}
\VoidTauInstructHeatCell{0}{3}{44.87}{39}{47}
\VoidTauInstructHeatCell{1}{3}{45.79}{39}{47}
\VoidTauInstructHeatCell{2}{3}{46.51}{39}{47}
\VoidTauInstructHeatCell{3}{3}{46.20}{39}{47}
\VoidTauInstructHeatCell{0}{2}{46.10}{39}{47}
\VoidTauInstructHeatCell{1}{2}{45.79}{39}{47}
\VoidTauInstructHeatCell{2}{2}{46.51}{39}{47}
\VoidTauInstructHeatCell{3}{2}{46.20}{39}{47}
\VoidTauInstructHeatCell{0}{1}{46.10}{39}{47}
\VoidTauInstructHeatCell{1}{1}{45.79}{39}{47}
\VoidTauInstructHeatCell{2}{1}{46.51}{39}{47}
\VoidTauInstructHeatCell{3}{1}{46.20}{39}{47}
\VoidTauInstructHeatCell{0}{0}{46.10}{39}{47}
\VoidTauInstructHeatCell{1}{0}{45.79}{39}{47}
\VoidTauInstructHeatCell{2}{0}{46.51}{39}{47}
\VoidTauInstructHeatCell{3}{0}{46.20}{39}{47}
\VoidTauInstructPanelFrame{MBPP}
\VoidTauInstructColorScale{39}{47}
\VoidTauInstructBaselineMark{39.73}{fixlen=64}{39}{47}{red!95!black}
\end{tikzpicture}
}
\end{subfigure}
\hfill
\begin{subfigure}[t]{0.19\textwidth}
\centering
\resizebox{\linewidth}{!}{
\begin{tikzpicture}
\VoidTauInstructHeatCell{0}{3}{51.44}{38}{52}
\VoidTauInstructHeatCell{1}{3}{51.21}{38}{52}
\VoidTauInstructHeatCell{2}{3}{51.78}{38}{52}
\VoidTauInstructHeatCell{3}{3}{51.16}{38}{52}
\VoidTauInstructHeatCell{0}{2}{51.75}{38}{52}
\VoidTauInstructHeatCell{1}{2}{51.21}{38}{52}
\VoidTauInstructHeatCell{2}{2}{51.78}{38}{52}
\VoidTauInstructHeatCell{3}{2}{51.16}{38}{52}
\VoidTauInstructHeatCell{0}{1}{51.75}{38}{52}
\VoidTauInstructHeatCell{1}{1}{51.21}{38}{52}
\VoidTauInstructHeatCell{2}{1}{51.78}{38}{52}
\VoidTauInstructHeatCell{3}{1}{51.16}{38}{52}
\VoidTauInstructHeatCell{0}{0}{51.75}{38}{52}
\VoidTauInstructHeatCell{1}{0}{51.21}{38}{52}
\VoidTauInstructHeatCell{2}{0}{51.78}{38}{52}
\VoidTauInstructHeatCell{3}{0}{51.16}{38}{52}
\VoidTauInstructPanelFrame{Mean}
\VoidTauInstructColorScale{38}{52}
\VoidTauInstructBaselineMark{38.99}{fixlen=64}{38}{52}{red!95!black}
\end{tikzpicture}
}
\end{subfigure}
\caption{VoidPadding-finetuned LLaDA-8B-Instruct + VoidExpansion threshold sweep. The sweep uses \(\tau_{\mathrm{nonvoid}}\in\{0.03,0.06,0.09,0.12\}\) and \(\tau_{\mathrm{gap}}\in\{-0.45,-0.40,-0.35,-0.30\}\); the blue box marks the selected setting from Table~\ref{tab:gen-length-robustness}, \((\tau_{\mathrm{nonvoid}},\tau_{\mathrm{gap}})=(0.09,-0.45)\). Cell color reports accuracy for reasoning tasks and pass@1 for code tasks. Red indicates lower scores and green indicates higher scores; the fixlen=64 markers are all red, showing that no-expansion fixed-64 is consistently low.}
\label{fig:void-expansion-threshold-ablation-instruct}
\end{figure*}
\begin{figure*}[t]
\centering
\newcommand{\VoidNFEHeatCell}[5]{
  \pgfmathsetmacro{\ratio}{max(0,min(1,(#3-#4)/(#5-#4)))}
  \pgfmathsetmacro{\pct}{100*\ratio}
  \pgfmathsetmacro{\bluepct}{22 + 40*\ratio}
  \path[draw=white, line width=0.35pt, fill=blue!\bluepct!white]
    (#1,#2) rectangle ++(1,1);
}
\newcommand{\VoidNFEPanelFrame}[1]{
  \node[font=\scriptsize\bfseries] at (2,4.74) {#1};
  \node[font=\tiny] at (2,4.38) {\(\tau_{\mathrm{nonvoid}}\)};
  \foreach \x/\lab in {0/0.03,1/0.06,2/0.09,3/0.12} {
    \node[font=\tiny] at (\x+0.5,4.14) {\lab};
  }
  \foreach \y/\lab in {3/{-0.45},2/{-0.40},1/{-0.35},0/{-0.30}} {
    \node[font=\tiny, anchor=east] at (-0.08,\y+0.5) {\lab};
  }
  \node[rotate=90,font=\tiny] at (-0.62,2) {\(\tau_{\mathrm{gap}}\)};
  \draw[black!55, line width=0.35pt] (0,0) rectangle (4,4);
  \draw[blue!80!black, line width=0.9pt] (2,3) rectangle ++(1,1);
}
\newcommand{\VoidNFEColorScale}[2]{
  \foreach \i in {0,...,19} {
    \pgfmathsetmacro{\bluepct}{22 + 40*\i/19}
    \path[draw=none, fill=blue!\bluepct!white] (4.12,{0.20*\i}) rectangle (4.48,{0.20*(\i+1)});
  }
  \draw[black!65, line width=0.45pt] (4.12,0) rectangle (4.48,4);
  \node[anchor=north,font=\scriptsize\bfseries] at (4.30,-0.08) {#1};
  \node[anchor=south,font=\scriptsize\bfseries] at (4.30,4.08) {#2};
}
\newcommand{\VoidNFEBBox}{\path[use as bounding box] (-0.85,-0.28) rectangle (5.85,5.00);}
\begin{minipage}[t]{\textwidth}
\centering
\begin{subfigure}[t]{0.19\linewidth}
\centering
\resizebox{\linewidth}{!}{
\begin{tikzpicture}
\VoidNFEBBox
\VoidNFEHeatCell{0}{3}{130.43}{130}{135}
\VoidNFEHeatCell{0}{2}{130.43}{130}{135}
\VoidNFEHeatCell{0}{1}{130.43}{130}{135}
\VoidNFEHeatCell{0}{0}{130.43}{130}{135}
\VoidNFEHeatCell{1}{3}{131.34}{130}{135}
\VoidNFEHeatCell{1}{2}{131.34}{130}{135}
\VoidNFEHeatCell{1}{1}{131.34}{130}{135}
\VoidNFEHeatCell{1}{0}{131.34}{130}{135}
\VoidNFEHeatCell{2}{3}{133.78}{130}{135}
\VoidNFEHeatCell{2}{2}{133.78}{130}{135}
\VoidNFEHeatCell{2}{1}{133.78}{130}{135}
\VoidNFEHeatCell{2}{0}{133.78}{130}{135}
\VoidNFEHeatCell{3}{3}{134.00}{130}{135}
\VoidNFEHeatCell{3}{2}{134.00}{130}{135}
\VoidNFEHeatCell{3}{1}{134.00}{130}{135}
\VoidNFEHeatCell{3}{0}{134.00}{130}{135}
\VoidNFEPanelFrame{GSM8K}
\VoidNFEColorScale{130}{135}
\end{tikzpicture}
}
\end{subfigure}
\hfill
\begin{subfigure}[t]{0.19\linewidth}
\centering
\resizebox{\linewidth}{!}{
\begin{tikzpicture}
\VoidNFEBBox
\VoidNFEHeatCell{0}{3}{418.05}{414}{419}
\VoidNFEHeatCell{0}{2}{418.05}{414}{419}
\VoidNFEHeatCell{0}{1}{418.05}{414}{419}
\VoidNFEHeatCell{0}{0}{418.05}{414}{419}
\VoidNFEHeatCell{1}{3}{418.05}{414}{419}
\VoidNFEHeatCell{1}{2}{418.05}{414}{419}
\VoidNFEHeatCell{1}{1}{418.05}{414}{419}
\VoidNFEHeatCell{1}{0}{418.05}{414}{419}
\VoidNFEHeatCell{2}{3}{418.05}{414}{419}
\VoidNFEHeatCell{2}{2}{418.05}{414}{419}
\VoidNFEHeatCell{2}{1}{418.05}{414}{419}
\VoidNFEHeatCell{2}{0}{418.05}{414}{419}
\VoidNFEHeatCell{3}{3}{414.39}{414}{419}
\VoidNFEHeatCell{3}{2}{414.39}{414}{419}
\VoidNFEHeatCell{3}{1}{414.39}{414}{419}
\VoidNFEHeatCell{3}{0}{414.39}{414}{419}
\VoidNFEPanelFrame{MATH500}
\VoidNFEColorScale{414}{419}
\end{tikzpicture}
}
\end{subfigure}
\hfill
\begin{subfigure}[t]{0.19\linewidth}
\centering
\resizebox{\linewidth}{!}{
\begin{tikzpicture}
\VoidNFEBBox
\VoidNFEHeatCell{0}{3}{88.27}{65}{89}
\VoidNFEHeatCell{0}{2}{88.27}{65}{89}
\VoidNFEHeatCell{0}{1}{88.27}{65}{89}
\VoidNFEHeatCell{0}{0}{88.27}{65}{89}
\VoidNFEHeatCell{1}{3}{77.88}{65}{89}
\VoidNFEHeatCell{1}{2}{77.88}{65}{89}
\VoidNFEHeatCell{1}{1}{77.88}{65}{89}
\VoidNFEHeatCell{1}{0}{77.88}{65}{89}
\VoidNFEHeatCell{2}{3}{73.06}{65}{89}
\VoidNFEHeatCell{2}{2}{73.06}{65}{89}
\VoidNFEHeatCell{2}{1}{73.06}{65}{89}
\VoidNFEHeatCell{2}{0}{73.06}{65}{89}
\VoidNFEHeatCell{3}{3}{65.75}{65}{89}
\VoidNFEHeatCell{3}{2}{65.75}{65}{89}
\VoidNFEHeatCell{3}{1}{65.75}{65}{89}
\VoidNFEHeatCell{3}{0}{65.75}{65}{89}
\VoidNFEPanelFrame{HumanEval}
\VoidNFEColorScale{65}{89}
\end{tikzpicture}
}
\end{subfigure}
\hfill
\begin{subfigure}[t]{0.19\linewidth}
\centering
\resizebox{\linewidth}{!}{
\begin{tikzpicture}
\VoidNFEBBox
\VoidNFEHeatCell{0}{3}{64.57}{63}{66}
\VoidNFEHeatCell{0}{2}{65.26}{63}{66}
\VoidNFEHeatCell{0}{1}{65.26}{63}{66}
\VoidNFEHeatCell{0}{0}{65.26}{63}{66}
\VoidNFEHeatCell{1}{3}{63.33}{63}{66}
\VoidNFEHeatCell{1}{2}{63.33}{63}{66}
\VoidNFEHeatCell{1}{1}{63.33}{63}{66}
\VoidNFEHeatCell{1}{0}{63.33}{63}{66}
\VoidNFEHeatCell{2}{3}{63.50}{63}{66}
\VoidNFEHeatCell{2}{2}{63.50}{63}{66}
\VoidNFEHeatCell{2}{1}{63.50}{63}{66}
\VoidNFEHeatCell{2}{0}{63.50}{63}{66}
\VoidNFEHeatCell{3}{3}{63.44}{63}{66}
\VoidNFEHeatCell{3}{2}{63.44}{63}{66}
\VoidNFEHeatCell{3}{1}{63.44}{63}{66}
\VoidNFEHeatCell{3}{0}{63.44}{63}{66}
\VoidNFEPanelFrame{MBPP}
\VoidNFEColorScale{63}{66}
\end{tikzpicture}
}
\end{subfigure}
\hfill
\begin{subfigure}[t]{0.19\linewidth}
\centering
\resizebox{\linewidth}{!}{
\begin{tikzpicture}
\VoidNFEBBox
\VoidNFEHeatCell{0}{3}{175.33}{169}{176}
\VoidNFEHeatCell{0}{2}{175.50}{169}{176}
\VoidNFEHeatCell{0}{1}{175.50}{169}{176}
\VoidNFEHeatCell{0}{0}{175.50}{169}{176}
\VoidNFEHeatCell{1}{3}{172.65}{169}{176}
\VoidNFEHeatCell{1}{2}{172.65}{169}{176}
\VoidNFEHeatCell{1}{1}{172.65}{169}{176}
\VoidNFEHeatCell{1}{0}{172.65}{169}{176}
\VoidNFEHeatCell{2}{3}{172.10}{169}{176}
\VoidNFEHeatCell{2}{2}{172.10}{169}{176}
\VoidNFEHeatCell{2}{1}{172.10}{169}{176}
\VoidNFEHeatCell{2}{0}{172.10}{169}{176}
\VoidNFEHeatCell{3}{3}{169.39}{169}{176}
\VoidNFEHeatCell{3}{2}{169.39}{169}{176}
\VoidNFEHeatCell{3}{1}{169.39}{169}{176}
\VoidNFEHeatCell{3}{0}{169.39}{169}{176}
\VoidNFEPanelFrame{Avg.}
\VoidNFEColorScale{169}{176}
\end{tikzpicture}
}
\end{subfigure}
\end{minipage}
\caption{VoidPadding-finetuned LLaDA-8B-Instruct + VoidExpansion threshold-sweep average NFE. The sweep uses \(\tau_{\mathrm{nonvoid}}\in\{0.03,0.06,0.09,0.12\}\) and \(\tau_{\mathrm{gap}}\in\{-0.45,-0.40,-0.35,-0.30\}\); the blue box marks the selected setting from Table~\ref{tab:gen-length-robustness}, \((\tau_{\mathrm{nonvoid}},\tau_{\mathrm{gap}})=(0.09,-0.45)\). Cell color reports benchmark-level average decode NFE for the first four panels, and Avg. is the arithmetic average over the four benchmarks.}
\label{fig:appendix-void-threshold-nfe}
\let\VoidNFEHeatCell\relax
\let\VoidNFEPanelFrame\relax
\let\VoidNFEColorScale\relax
\let\VoidNFEBBox\relax
\end{figure*}
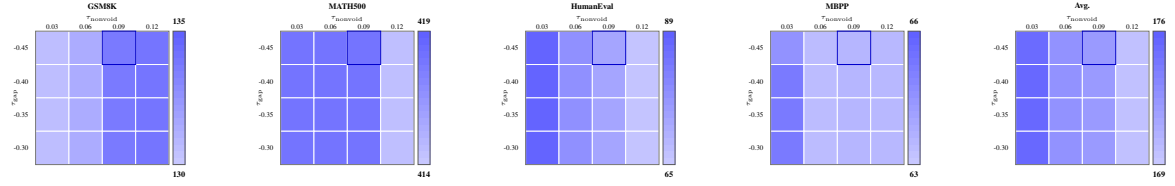
\begin{figure*}[t]
\centering
\newcommand{\VoidTauBBox}{\path[use as bounding box] (-0.85,-0.28) rectangle (5.85,5.00);}
\begin{minipage}[t]{\textwidth}
\centering
\begin{subfigure}[t]{0.19\linewidth}
\centering
\resizebox{\linewidth}{!}{
\begin{tikzpicture}
\VoidTauBBox
\VoidTauHeatCell{0}{3}{76.57}{60}{77}
\VoidTauHeatCell{1}{3}{76.72}{60}{77}
\VoidTauHeatCell{2}{3}{75.97}{60}{77}
\VoidTauHeatCell{3}{3}{75.36}{60}{77}
\VoidTauHeatCell{0}{2}{76.57}{60}{77}
\VoidTauHeatCell{1}{2}{76.72}{60}{77}
\VoidTauHeatCell{2}{2}{75.97}{60}{77}
\VoidTauHeatCell{3}{2}{75.36}{60}{77}
\VoidTauHeatCell{0}{1}{76.57}{60}{77}
\VoidTauHeatCell{1}{1}{76.72}{60}{77}
\VoidTauHeatCell{2}{1}{75.97}{60}{77}
\VoidTauHeatCell{3}{1}{75.36}{60}{77}
\VoidTauHeatCell{0}{0}{76.57}{60}{77}
\VoidTauHeatCell{1}{0}{76.72}{60}{77}
\VoidTauHeatCell{2}{0}{75.97}{60}{77}
\VoidTauHeatCell{3}{0}{75.36}{60}{77}
\VoidTauPanelFrame{GSM8K}
\VoidTauColorScale{60}{77}
\VoidTauBaselineMark{60.12}{fixlen=64}{60}{77}{red!95!black}
\end{tikzpicture}
}
\end{subfigure}
\hfill
\begin{subfigure}[t]{0.19\linewidth}
\centering
\resizebox{\linewidth}{!}{
\begin{tikzpicture}
\VoidTauBBox
\VoidTauHeatCell{0}{3}{35.40}{27}{37}
\VoidTauHeatCell{1}{3}{37.00}{27}{37}
\VoidTauHeatCell{2}{3}{35.20}{27}{37}
\VoidTauHeatCell{3}{3}{34.40}{27}{37}
\VoidTauHeatCell{0}{2}{35.40}{27}{37}
\VoidTauHeatCell{1}{2}{37.00}{27}{37}
\VoidTauHeatCell{2}{2}{34.80}{27}{37}
\VoidTauHeatCell{3}{2}{33.20}{27}{37}
\VoidTauHeatCell{0}{1}{35.40}{27}{37}
\VoidTauHeatCell{1}{1}{37.00}{27}{37}
\VoidTauHeatCell{2}{1}{35.20}{27}{37}
\VoidTauHeatCell{3}{1}{34.40}{27}{37}
\VoidTauHeatCell{0}{0}{35.40}{27}{37}
\VoidTauHeatCell{1}{0}{37.00}{27}{37}
\VoidTauHeatCell{2}{0}{35.20}{27}{37}
\VoidTauHeatCell{3}{0}{34.40}{27}{37}
\VoidTauPanelFrame{MATH500}
\VoidTauColorScale{27}{37}
\VoidTauBaselineMark{27.80}{fixlen=64}{27}{37}{red!95!black}
\end{tikzpicture}
}
\end{subfigure}
\hfill
\begin{subfigure}[t]{0.19\linewidth}
\centering
\resizebox{\linewidth}{!}{
\begin{tikzpicture}
\VoidTauBBox
\VoidTauHeatCell{0}{3}{39.02}{34}{40}
\VoidTauHeatCell{1}{3}{39.63}{34}{40}
\VoidTauHeatCell{2}{3}{39.02}{34}{40}
\VoidTauHeatCell{3}{3}{37.20}{34}{40}
\VoidTauHeatCell{0}{2}{39.02}{34}{40}
\VoidTauHeatCell{1}{2}{39.63}{34}{40}
\VoidTauHeatCell{2}{2}{39.02}{34}{40}
\VoidTauHeatCell{3}{2}{37.20}{34}{40}
\VoidTauHeatCell{0}{1}{39.02}{34}{40}
\VoidTauHeatCell{1}{1}{39.63}{34}{40}
\VoidTauHeatCell{2}{1}{39.02}{34}{40}
\VoidTauHeatCell{3}{1}{37.20}{34}{40}
\VoidTauHeatCell{0}{0}{39.02}{34}{40}
\VoidTauHeatCell{1}{0}{39.63}{34}{40}
\VoidTauHeatCell{2}{0}{39.02}{34}{40}
\VoidTauHeatCell{3}{0}{37.20}{34}{40}
\VoidTauPanelFrame{HumanEval}
\VoidTauColorScale{34}{40}
\VoidTauBaselineMark{34.76}{fixlen=64}{34}{40}{red!95!black}
\end{tikzpicture}
}
\end{subfigure}
\hfill
\begin{subfigure}[t]{0.19\linewidth}
\centering
\resizebox{\linewidth}{!}{
\begin{tikzpicture}
\VoidTauBBox
\VoidTauHeatCell{0}{3}{44.66}{41}{46}
\VoidTauHeatCell{1}{3}{44.46}{41}{46}
\VoidTauHeatCell{2}{3}{44.66}{41}{46}
\VoidTauHeatCell{3}{3}{44.87}{41}{46}
\VoidTauHeatCell{0}{2}{44.66}{41}{46}
\VoidTauHeatCell{1}{2}{44.46}{41}{46}
\VoidTauHeatCell{2}{2}{44.66}{41}{46}
\VoidTauHeatCell{3}{2}{44.87}{41}{46}
\VoidTauHeatCell{0}{1}{44.66}{41}{46}
\VoidTauHeatCell{1}{1}{44.46}{41}{46}
\VoidTauHeatCell{2}{1}{44.66}{41}{46}
\VoidTauHeatCell{3}{1}{44.87}{41}{46}
\VoidTauHeatCell{0}{0}{44.66}{41}{46}
\VoidTauHeatCell{1}{0}{44.46}{41}{46}
\VoidTauHeatCell{2}{0}{44.66}{41}{46}
\VoidTauHeatCell{3}{0}{44.87}{41}{46}
\VoidTauPanelFrame{MBPP}
\VoidTauColorScale{41}{46}
\VoidTauBaselineMark{41.17}{fixlen=64}{41}{46}{red!95!black}
\end{tikzpicture}
}
\end{subfigure}
\hfill
\begin{subfigure}[t]{0.19\linewidth}
\centering
\resizebox{\linewidth}{!}{
\begin{tikzpicture}
\VoidTauBBox
\VoidTauHeatCell{0}{3}{48.91}{40}{50}
\VoidTauHeatCell{1}{3}{49.45}{40}{50}
\VoidTauHeatCell{2}{3}{48.71}{40}{50}
\VoidTauHeatCell{3}{3}{47.96}{40}{50}
\VoidTauHeatCell{0}{2}{48.91}{40}{50}
\VoidTauHeatCell{1}{2}{49.45}{40}{50}
\VoidTauHeatCell{2}{2}{48.61}{40}{50}
\VoidTauHeatCell{3}{2}{47.66}{40}{50}
\VoidTauHeatCell{0}{1}{48.91}{40}{50}
\VoidTauHeatCell{1}{1}{49.45}{40}{50}
\VoidTauHeatCell{2}{1}{48.71}{40}{50}
\VoidTauHeatCell{3}{1}{47.96}{40}{50}
\VoidTauHeatCell{0}{0}{48.91}{40}{50}
\VoidTauHeatCell{1}{0}{49.45}{40}{50}
\VoidTauHeatCell{2}{0}{48.71}{40}{50}
\VoidTauHeatCell{3}{0}{47.96}{40}{50}
\VoidTauPanelFrame{Mean}
\VoidTauColorScale{40}{50}
\VoidTauBaselineMark{40.96}{fixlen=64}{40}{50}{red!95!black}
\end{tikzpicture}
}
\end{subfigure}
\caption{VoidPadding-finetuned LLaDA-8B-Base + VoidExpansion threshold sweep. The sweep uses \(\tau_{\mathrm{nonvoid}}\in\{0.06,0.09,0.12,0.15\}\) and \(\tau_{\mathrm{gap}}\in\{-0.55,-0.50,-0.45,-0.40\}\); the blue box marks the selected setting, \((\tau_{\mathrm{nonvoid}},\tau_{\mathrm{gap}})=(0.09,-0.45)\). Cell color reports accuracy for reasoning tasks and pass@1 for code tasks. Red indicates lower scores and green indicates higher scores; the fixlen=64 markers are all red, showing that no-expansion fixed-64 is consistently low.}
\label{fig:void-expansion-threshold-ablation}
\end{minipage}
\let\VoidTauBBox\relax
\end{figure*}
\input{figures/dream_void_expansion_threshold_ablation}

\end{document}